\documentclass[twoside,11pt]{article}
\usepackage[preprint]{jmlr2e}
\usepackage{blindtext}

\usepackage{jmlr2e}
\usepackage{amsmath,amsfonts}
\usepackage{algorithmicx}
\usepackage{algorithm}
\usepackage{algpseudocode}

\usepackage{array}

\usepackage{textcomp}
\usepackage{stfloats}
\usepackage{url}
\usepackage{verbatim}
\usepackage{graphicx}
\usepackage{balance}
\usepackage{color}
\usepackage{todonotes}
\usepackage{bm}
\usepackage{amssymb}
\usepackage{booktabs}
\usepackage{multirow}
\usepackage{siunitx}
\usepackage{subfigure}
\usepackage{float}
\usepackage{threeparttable}
\usepackage{enumitem}
\firstpageno{1}

\usepackage{lastpage}

\jmlrheading{23}{2022}{1-\pageref{LastPage}}{1/21; Revised 5/22}{9/22}{21-0000}{%
    Ziming Wang, 
    Changwu Huang, 
    Ke Tang, and 
    Xin Yao%
}

\ShortHeadings{Procedural Fairness and Its Relationship with Distributive Fairness}{Wang, Huang, Tang, Yao}
\firstpageno{1}

\begin{document}

\title{Procedural Fairness and Its Relationship with Distributive Fairness in Machine Learning}

\author{\name Ziming Wang \email wangzm2021@mail.sustech.edu.cn \\
       \addr Guangdong Provincial Key Laboratory of Brain-inspired Intelligent Computation\\ Department of Computer Science and Engineering\\ Southern University of Science and Technology\\ Shenzhen, 518055, China
       \AND
       \name Changwu Huang\thanks{Corresponding authors.} \email huangcw3@sustech.edu.cn \\
       \addr Guangdong Provincial Key Laboratory of Brain-inspired Intelligent Computation\\ Department of Computer Science and Engineering\\ Southern University of Science and Technology\\
       Shenzhen, 518055, China
       \AND
       \name Ke Tang \email tangk3@sustech.edu.cn \\
       \addr  Guangdong Provincial Key Laboratory of Brain-inspired Intelligent Computation\\ Department of Computer Science and Engineering\\ Southern University of Science and Technology\\
       Shenzhen, 518055, China
       \AND
       \name Xin Yao\footnotemark[1] \email xinyao@ln.edu.hk \\
       \addr Department of Computer Science and Engineering\\ Southern University of Science and Technology\\
       Shenzhen, 518055, China\\
       \addr School of Data Science\\ Lingnan University\\
       Hong Kong SAR
       }

\maketitle

\begin{abstract}
Fairness in machine learning (ML) has garnered significant attention in recent years. While existing research has predominantly focused on the distributive fairness of ML models, there has been limited exploration of procedural fairness. This paper proposes a novel method to achieve procedural fairness during the model training phase. The effectiveness of the proposed method is validated through experiments conducted on one synthetic and six real-world datasets. Additionally, this work studies the relationship between procedural fairness and distributive fairness in ML models. On one hand, the impact of dataset bias and the procedural fairness of ML model on its distributive fairness is examined. The results highlight a significant influence of both dataset bias and procedural fairness on distributive fairness. On the other hand, the distinctions between optimizing procedural and distributive fairness metrics are analyzed. Experimental results demonstrate that optimizing procedural fairness metrics mitigates biases introduced or amplified by the decision-making process, thereby ensuring fairness in the decision-making process itself, as well as improving distributive fairness. In contrast, optimizing distributive fairness metrics encourages the ML model's decision-making process to favor disadvantaged groups, counterbalancing the inherent preferences for advantaged groups present in the dataset and ultimately achieving distributive fairness.
\end{abstract}


\section{Introduction}
\label{sec:introduction} 

With the wide application of artificial intelligence (AI), the ethical considerations surrounding AI have attracted increasing attention~\citep{9844014,li2021trust}. In particular, the issue of fairness in AI has emerged as a paramount concern~\citep{mehrabi2021survey,pessach2022review,caton2020fairness}. Ensuring fairness of AI-based decision-making becomes crucial to avoid exacerbating existing social inequalities and promoting equitable outcomes for individuals and communities. Therefore, research on fairness in AI and machine learning (ML) becomes imperative to foster responsible and inclusive deployment of AI technologies.

Fairness in decision-making involves two fundamental aspects: distributive fairness (a.k.a outcome fairness), which focuses on the fairness of decision outcomes~\citep{greenberg1987taxonomy,morse2021ends,grgic2018beyond}, and procedural fairness (a.k.a process fairness), which examines the fairness of the decision-making process~\citep{morse2021ends,grgic2018beyond}. While research on AI fairness has predominantly concentrated on distributive fairness~\citep{mehrabi2021survey,pessach2022review}, there is increasing recognition of the importance of procedural fairness~\citep{grgic2018beyond}. To simplify the discussion, we will use the term ``distributively-fair ML model" to refer to an ML model that meets some distributive fairness criteria, while a ``distributively-unfair ML model" refers to an ML model that does not meet these criteria. Similarly, a ``procedurally-fair ML model" indicates an ML model that aligns with certain procedural fairness metrics, whereas a ``procedurally-unfair ML model" is one that does not meet the procedural fairness criterion.

In the limited research on procedural fairness,~\cite{grgic2018beyond} assessed the procedural fairness of the ML model based on human judgments of the fairness of the input features. However, this method could not truly capture the fairness of the ML model's decision-making process. Moreover, the human judgment-based approach limits its wide application and makes it difficult to ensure its objectivity~\citep{wang2024procedural}. Therefore,~\cite{wang2024procedural} utilized explainable artificial intelligence (XAI) techniques~\citep{wang2024roadmap} to capture the decision-making process of the ML model and proposed a quantitative metric, called \textit{GPF\textsubscript{FAE}}, for assessing the model's group procedural fairness.

Although procedural fairness was considered crucial for achieving overall fairness and was viewed as a more reliable assessment criterion than distributive fairness~\citep{morse2021ends,thibaut1975procedural,lind2001heuristic,van1998we}, there is a scarcity of methods available to ensure procedural fairness. Previous approaches have included feature selection based on human judgment to attain procedurally-fair ML models~\citep{grgic2018beyond}, as well as the use of XAI technology to enhance procedural fairness in unfair ML models~\citep{wang2024procedural}. In general, there is still a lack of methods for learning procedurally-fair ML models during the training process. This motivates us to consider the procedural fairness metric \textit{GPF\textsubscript{FAE}}~\citep{wang2024procedural} when constructing ML models, thereby introducing a novel method to achieve procedural fairness during the ML model training process. 

Moreover, biases (or unfairness) could appear in a training dataset, in model's decision process or in model outcomes. It has been rather unclear in the existing literature about the relationships among the three and lacks a corresponding analysis. This paper fills in this gap in the literature. On one hand, we demonstrate that the model's distributive unfairness stems from a combination of inherent biases in the dataset and procedural unfairness within the ML model, and the unfairness of the two can also be superimposed and canceled out. This analysis helps to identify the sources of distributive unfairness. On the other hand, we study quantitative metrics for assessing procedural and distributive fairness, along with corresponding optimization methods. By leveraging these metrics and methodologies, we analyze the distinctions between these two dimensions of fairness in ML. Our work could provide a valuable guidance to stakeholders in selecting appropriate fairness metrics and methodologies. To summarize, this paper addresses the following two research questions (RQs):

\begin{enumerate}[leftmargin=27pt]
    \item[\textbf{\textit{RQ1}}] What are the influences of inherent dataset bias and the ML model's procedural fairness on its distributive fairness?
    \item[\textbf{\textit{RQ2}}] What are the differences between optimizing procedural fairness metrics and distributive fairness metrics in ML?
\end{enumerate}

In this paper, we first propose a method to achieve procedural fairness during training and then answer the above two research questions. The main contributions of this paper are summarized as follows:

\begin{enumerate}
    \item[(1)] A novel method to construct procedurally-fair ML models that optimizes the metric \textit{GPF\textsubscript{FAE}} during training phase is proposed. The effectiveness of the proposed method is validated through comprehensive experiments on both synthetic and real-world datasets. 
    \item[(2)] The impact of both the inherent bias of the dataset and the procedural fairness of the ML model on the ML model's distributive fairness is analyzed. Experimental results demonstrate that when the dataset is unbiased and the ML model is procedurally-fair, it is also distributively-fair. Conversely, the presence of either dataset bias or procedural unfairness in the ML model leads to distributive unfairness, with the procedural fairness having a stronger influence. Additionally, when both dataset bias and unfairness in the ML model's decision process are present, their impacts on distributive fairness could superimpose or cancel each other out depending on whether the two are aligned or opposite. 
    \item[(3)] The disparities between considering procedural and distributive fairness metrics in ML are examined and revealed. When the dataset is biased, optimizing procedural fairness metrics will rectify or mitigate the bias introduced or magnified by the ML model, thus achieving procedural fairness and improving distributive fairness. In contrast, optimizing distributive fairness metrics results in an ML model where the decision process favors the disadvantaged group, counterbalancing the inherent preference for the advantaged group present in the dataset, thereby achieving distributive fairness. In other words, training an ML model with unfair decision-making process on a biased training dataset can achieve distributive fairness.
\end{enumerate}

The rest of this paper is organized as follows. Section \ref{sec:related_work} presents the related work on procedural and distributive fairness in ML. In Section \ref{sec:method}, we propose an in-process method to achieve procedural fairness and validate its effectiveness. Section \ref{sec:relationship} examines the relationship between procedural fairness and distributive fairness, thus answering the two research questions. Further discussions are provided in Section \ref{sec:discussion}. Section \ref{sec:conclusion} concludes the paper and points out some future research directions.

\section{Related Work}
\label{sec:related_work}

In this section, we provide a brief overview of related work on fairness in ML. We begin by presenting existing research on procedural fairness within ML models, with a specific focus on the \textit{GPF\textsubscript{FAE}}, a metric for evaluating group procedural fairness proposed by~\cite{wang2024procedural}. Subsequently, we present evaluation metrics and improvement methods for the ML model's distributive fairness, which have been studied extensively in the literature.

\subsection{Existing Research on Procedural Fairness in ML}

Currently, procedural fairness is still relatively understudied in the field of ML. Among the existing research,~\cite{grgic2018beyond} conducted an assessment of procedural fairness by relying on human judgments concerning the fairness of input features. In addition, they used feature selection to remove features deemed unfair by manual evaluation, thus improving the procedural fairness of the ML model~\citep{grgic2018beyond}. However, this method of assessing the procedural fairness of ML models does not really consider the decision-making process or logic of the ML model, and its assessment results are susceptible to human subjectivity~\citep{wang2024procedural}. 

As an alternative,~\cite{wang2024procedural} first defined group procedural fairness as “\textit{similar data points in two groups should have similar decision process or logic}". Subsequently, it proposed the metric \textit{GPF\textsubscript{FAE}} for assessing the group procedural fairness of ML models. \textit{GPF\textsubscript{FAE}} utilizes the feature attribution explanation (FAE) method to portray the decision logic of the ML model. The FAE method, widely acknowledged in the realm of XAI, elucidates the decision-making processes of ML models by quantifying the attribution of each input feature for a given data point~\citep{10.1145/3617380}. Depending on how the explanations are generated, the FAE method can be categorized into perturbation-based methods (e.g., \textit{SHAP}~\citep{lundberg2017unified}) and gradient-based methods (e.g., \textit{Grad}~\citep{simonyan2014deep}). Among them, gradient-based is computationally efficient, while perturbation-based is more theoretically secure but computationally expensive. And~\cite{wang2024procedural} employed \textit{SHAP}~\citep{lundberg2017unified} as the chosen FAE method. Specifically, \textit{GPF\textsubscript{FAE}} is defined as~\citep{wang2024procedural}: 

\begin{equation}
\begin{split}
    \textit{GPF\textsubscript{FAE}}=&d(\textit{\textbf{E}}_1,\textit{\textbf{E}}_2);\\
    \textit{\textbf{E}}_1=&\{\textit{\textbf{e}}_1^{(i)}|\textit{\textbf{e}}_1^{(i)}=g(f_{\bm{\theta}}, \textit{\textbf{x}}_1^{(i)}),\textit{\textbf{x}}_1^{(i)}\in \textit{\textbf{X}}^{'}_{1}\},\\
    \textit{\textbf{E}}_2=&\{\textit{\textbf{e}}_2^{(i)}|\textit{\textbf{e}}_2^{(i)}=g(f_{\bm{\theta}}, \textit{\textbf{x}}_2^{(i)}),\textit{\textbf{x}}_2^{(i)}\in \textit{\textbf{X}}^{'}_{2}\}, \\
\end{split}
\end{equation}

\noindent where $\textit{\textbf{X}}^{'}_{1}=\{\textit{\textbf{x}}_1^{(1)}, \dots, \textit{\textbf{x}}_1^{(n)}\}$ and $\textit{\textbf{X}}^{'}_{2}=\{\textit{\textbf{x}}_2^{(1)}, \dots, \textit{\textbf{x}}_2^{(n)}\}$ represent sets of $n$ data points selected from two distinct groups $\textit{\textbf{X}}_{1}$ and $\textit{\textbf{X}}_{2}$, such as male and female groups, and each pair $\textit{\textbf{x}}_1^{(i)}\in \textit{\textbf{X}}^{'}_{1}$ and $\textit{\textbf{x}}_2^{(i)}\in \textit{\textbf{X}}^{'}_{2}$ are the most similar data points to each other in different groups. $g$ denotes a local FAE function that takes an explained ML model $f_{\bm{\theta}}$ and an explained data point $\textit{\textbf{x}}^{(i)}$ as inputs and returns FAE explanations (i.e., feature importance scores) $\textit{\textbf{e}}^{(i)}=g(f_{\bm{\theta}},\,\textit{\textbf{x}}^{(i)})\in\mathbb{R}^{d}$. $d(\cdot, \cdot)$ is a measurement of the distance between two sets of FAE explanation results $\textit{\textbf{E}}_1$ and $\textit{\textbf{E}}_2$. Consistent with~\cite{wang2024procedural}, we used the maximum mean discrepancy (MMD) with an exponential kernel function to measure the difference between the distributions of the two explanation sets. The $p$-value obtained by performing a permutation test on the generated kernel matrix is used as the final evaluation result.

Straightforwardly, the \textit{GPF\textsubscript{FAE}} metric utilizes \textit{SHAP}~\citep{lundberg2017unified} to portray the decision logic of the ML model $f_{\bm{\theta}}$. It evaluates the difference in the explanations of similar data points in two groups to assess the procedural fairness of $f_{\bm{\theta}}$. The closer the \textit{GPF\textsubscript{FAE}} metric is to 1.0, the more similar the explanation results of similar data points in the two groups are, and the fairer the decision-making process of the ML model.

Although~\cite{wang2024procedural} proposed two methods to improve the procedural fairness of an already trained ML model, there is a lack of methods for learning procedurally-fair ML models during the training process. Therefore, this paper proposes the first method to achieve procedural fairness during training by considering the metric \textit{GPF\textsubscript{FAE}}.

\subsection{Existing Metrics and Methods for Distributive Fairness}

In recent years, distributive fairness in ML has received a lot of attention and been extensively studied~\citep{pessach2022review}. In this context, we focus on group distributive fairness, providing a brief review of its metrics and methods. Group distributive fairness refers to the treatment of different group equity by ML models~\citep{mehrabi2021survey}. To assess and quantify this aspect, researchers have introduced various evaluation metrics. Among these metrics, one of the most widely used criteria revolves around the notion that the proportion of positive outcomes generated for two distinct groups should be close to each other. Depending on whether this concept is expressed as a ratio or a difference, it is defined as disparate impact (DI)~\citep{feldman2015certifying} and demographic parity (DP)~\citep{dwork2012fairness,calders2010three}, as shown below:

\begin{equation}
\label{eq:DI}
DI = \frac{P\left(\hat{y}=1 | s=s_{1}\right)}{P\left(\hat{y}=1 | s=s_{2}\right)},
\end{equation}

\begin{equation}
\label{eq:DP}
DP = \left | P\left(\hat{y}=1 | s=s_{1}\right) - P\left(\hat{y}=1 | s=s_{2}\right) \right |,
\end{equation}

\noindent where $\hat{y}$ is the predicted labels of the ML model $f_{\bm{\theta}}$ parameterized by $\bm{\theta}$, $s$ represents the protected attribute (e.g., sex), and $s_1$ and $s_2$ denote two different groups associated with the sensitive attribute.

Moreover, it is worth noting that various legal regulations, both explicitly and implicitly, incorporate requirements related to \textit{DI} and \textit{DP} metrics. For example, the Uniform Guidelines on Employee Selection Procedures in the~\cite{equal1990uniform} require that the selection rate for any race, gender, or ethnic group should not be less than four-fifths of the highest group. In addition,~\cite{wachter2020bias} pointed out that some of the tests employed by the European Court of Justice and Member State courts to measure indirect discrimination align with the \textit{DP} metric in algorithmic fairness. Therefore, in this paper, we use the \textit{DP} metric to assess the ML model's distributive fairness.

Based on the defined distributive fairness metrics, researchers have introduced various in-process techniques aimed at integrating distributive fairness considerations into the training of ML algorithms~\citep{mehrabi2021survey}, including regularization-based~\citep{kamishima2012fairness}, constraint-based~\citep{agarwal2018reductions}, and adversarial learning-based~\citep{zhang2018mitigating}. These techniques are primarily employed to mitigate bias within ML models. In this paper, we use the three aforementioned methodologies to optimize the distributive fairness metric \textit{DP}. Our goal is to elucidate the distinctions between optimizing distributive fairness metrics and procedural fairness metrics in ML.

Furthermore,~\cite{zemel2013learning} proposed a pre-process method called learning fair representations (LFR). It aims to find good representations of the data with two goals: to encode the data as well as possible while obfuscating any information related to the sensitive attribute~\citep{zemel2013learning}. This approach effectively learns fair representations of data, thereby improving the distributive fairness of ML models. In this paper, we utilize LFR to procure unbiased datasets to assist in exploring the relationship between data bias, and unfairness in the decision-making process and outcomes of ML models.

\section{Achieving Procedural Fairness in ML}
\label{sec:method}

In this section, we propose the first method to achieve procedural fairness during the training process by optimizing the \textit{GPF\textsubscript{FAE}} metric and validate its effectiveness on one synthetic and six real-world datasets. The code used in this paper is available at \url{https://github.com/oddwang/Procedural-Fairness-Relationship}.

\subsection{Notations}

We focus on a binary classification problem aiming at learning a mapping function between input feature vectors $\textit{\textbf{x}}\in\mathbb{R}^{d}$ and class label ${y}\in \{0, 1\}$. The goal is to find a classifier $f_{\bm{\theta}}: \textit{\textbf{x}} \mapsto {y}$ based on the dataset $\bm{\mathcal{D}}=(\textit{\textbf{X}},\textit{\textbf{Y}})= \{ (\textit{\textbf{x}}^{(i)},y^{(i)})\}_{i=1}^{m}$, where $\textit{\textbf{x}}^{(i)}=[ x_{1}^{(i)},\cdots, x_{d}^{(i)}] \in  \mathbb{R}^{d} $ are feature vectors and $y^{(i)} \in \{0, 1\}$ are their corresponding label, such that given a feature vector $\textit{\textbf{x}}$ with unknown label $y$, the classifier predicts its label $\hat{y}=f_{\bm{\theta}}\left (\textit{\textbf{x}}\right )$. In addition, each data point $\textit{\textbf{x}}$ has a sensitive attribute $s$ (e.g., sex), which indicates the data point's group membership, and the ML model $f_{\bm{\theta}}$ needs to be fair to the sensitive attribute. There can be more than one sensitive attribute, but in this paper, without loss of generality, we consider the case of one sensitive attribute (e.g., the gender of each data point $s=\{male, female\}$) and denote the two different values of the sensitive attribute by $s_1$ and $s_2$. Each data point $(\textit{\textbf{x}}^{(i)},y^{(i)})\in\bm{\mathcal{D}}$ has an associate sensitive feature value $s^{(i)}\in\{s_1,s_2\}$. In addition, the subsets of dataset $\bm{\mathcal{D}}$ with values $s=s_1$ and $s=s_2$ are denoted as $\bm{\mathcal{D}}_1=\{(\textit{\textbf{X}}_1,\textit{\textbf{Y}}_1)\in\bm{\mathcal{D}}\}=\{(\textit{\textbf{x}}^{(i)},y^{(i)})\in\bm{\mathcal{D}}|s^{(i)}=s_1\}$ and $\bm{\mathcal{D}}_2=\{(\textit{\textbf{X}}_2,\textit{\textbf{Y}}_2)\in\bm{\mathcal{D}}\}=\{(\textit{\textbf{x}}^{(i)},y^{(i)})\in\bm{\mathcal{D}}|s^{(i)}=s_2\}$, respectively. We use $g$ to represent a local FAE function, which takes an explained ML model $f_{\bm{\theta}}$ and an explained data point $\textit{\textbf{x}}^{(i)}$ as inputs and returns FAE explanations (i.e., feature importance scores) $\textit{\textbf{e}}^{(i)}=g(f_{\bm{\theta}},\,\textit{\textbf{x}}^{(i)})\in\mathbb{R}^{d}$.

\subsection{Achieving Procedural Fairness by Optimizing \textit{GPF\textsubscript{FAE}}}

We consider the \textit{GPF\textsubscript{FAE}} metric~\citep{wang2024procedural} in the form of a regularization term, thus optimizing the procedural fairness metric of the ML model during training. However, directly optimizing the \textit{GPF\textsubscript{FAE}} metric is undesirable and suffers from the following two major problems: (1) the $\textit{GPF\textsubscript{FAE}}$ metric is non-differentiable~\citep{wang2024procedural}, rendering it unsuitable for optimization within regular ML models; (2) optimizing the \textit{GPF\textsubscript{FAE}} metric during training necessitates the acquisition of explanation results from the current ML model at each epoch. Regrettably, the \textit{SHAP} explanation method~\citep{lundberg2017unified} is notably time-consuming, rendering it not suitable for training purposes.

To address problem (1), in our proposed method, we incorporate a revised form of the \textit{GPF\textsubscript{FAE}} metric as a penalty term in the loss function, denoted as the procedural fairness loss $\mathcal{L}_{GPF}$, during training the ML model. Specifically, we utilized the $\ell_1$ distance for evaluating the differences between the explanations of similar data points in two explanation sets as follows:
\begin{equation}
\label{eq:l1-gpf}
\mathcal{L}_{GPF} =\frac{1}{k}\sum _{i=1}^k\ell_1(\textit{\textbf{e}}_1^{(i)},\textit{\textbf{e}}_2^{(i)})=\frac{1}{k}\sum _{i=1}^k \left \| \textit{\textbf{e}}_1^{(i)}-\textit{\textbf{e}}_2^{(i)} \right \|_1 .
\end{equation}

To overcome problem (2), i.e., the substantial time cost associated with the \textit{SHAP} method~\citep{lundberg2017unified}, we replaced it with the gradient-based FAE method \textit{Grad}~\citep{simonyan2014deep} in the training phase to obtain explanations during the training process. It is worth noting that the computational efficiency inherent to \textit{Grad}~\citep{simonyan2014deep} eliminated the necessity of selecting $n$ pairs of similar data points from the datasets $\textit{\textbf{X}}_{1}$ and $\textit{\textbf{X}}_{2}$ to acquire their explanations $\textit{\textbf{E}}_1$ and $\textit{\textbf{E}}_2$. Instead, we were able to obtain explanations for all data points alongside their corresponding similar data points. In other words, the parameter $k$ in Eq. (\ref{eq:l1-gpf}) is the number of data points in the training set. $\mathcal{L}_{GPF}$ calculates the average of the differences in the explanations obtained on the \textit{Grad} method~\citep{simonyan2014deep} for all data points in the training set and their similar data points. The pseudo-code of our method is shown in Algorithm \ref{alg:method}.

\begin{algorithm}[t]
	\caption{Procedurally fair machine learning: Optimizing \textit{GPF\textsubscript{FAE}} to achieve procedural fairness.} 
    \label{alg:method}
	\begin{algorithmic}[1]
        \Require A randomly initialized ML model $f_{\bm{\theta}}$, training dataset $\bm{\mathcal{D}}=(\textit{\textbf{X}},\textit{\textbf{Y}})$, the data point similarity metric $d_x$
        \Ensure The trained ML model $f_{\bm{\theta}}$
        \State $\textit{\textbf{X}}_{1}=\{\textit{\textbf{x}}^{(i)}\in \textit{\textbf{X}}|s^{(i)}=s_1\}$
        \State $\textit{\textbf{X}}_{2}=\{\textit{\textbf{x}}^{(i)}\in \textit{\textbf{X}}|s^{(i)}=s_2\}$
        \State $\textit{\textbf{X}}^{'}_1=\emptyset, \textit{\textbf{X}}^{'}_2=\emptyset$
        \For{$\textbf{\textit{x}}^{(i)}_1$ in $\textit{\textbf{X}}_{1}$}
            \State Find the $\textbf{\textit{x}}^{(j)}_2\in \textit{\textbf{X}}_{2}$ with the smallest $d_x(\textbf{\textit{x}}^{(i)}_1,\textbf{\textit{x}}^{(j)}_2)$
            \State $\textit{\textbf{X}}^{'}_1=\textit{\textbf{X}}^{'}_1\cup\textbf{\textit{x}}^{(i)}_1$, $\textit{\textbf{X}}^{'}_2=\textit{\textbf{X}}^{'}_2\cup\textbf{\textit{x}}^{(j)}_2$
        \EndFor
        \For{$\textbf{\textit{x}}^{(j)}_2$ in $\textit{\textbf{X}}_{2}$}
            \State Find the $\textbf{\textit{x}}^{(i)}_1\in \textit{\textbf{X}}_{1}$ with the smallest $d_x(\textbf{\textit{x}}^{(j)}_2,\textbf{\textit{x}}^{(i)}_1)$
            \State $\textit{\textbf{X}}^{'}_1=\textit{\textbf{X}}^{'}_1\cup\textbf{\textit{x}}^{(i)}_1$, $\textit{\textbf{X}}^{'}_2=\textit{\textbf{X}}^{'}_2\cup\textbf{\textit{x}}^{(j)}_2$
        \EndFor 
        \While{iteration termination conditions are not reached}
            \State Calculate cross entropy loss $\mathcal{L}_{CE}$ with respect to $f_{\bm{\theta}}$
            \State $\textit{\textbf{E}}^{'}_1=\{\textit{\textbf{e}}_1^{(i)}|\textit{\textbf{e}}_1^{(i)}\,=g(f_{\bm{\theta}}, \textit{\textbf{x}}_1^{(i)}\,),\textit{\textbf{x}}_1^{(i)}\in \textit{\textbf{X}}^{'}_1\}$
            \State $\textit{\textbf{E}}^{'}_2=\{\textit{\textbf{e}}_2^{(i)}|\textit{\textbf{e}}_2^{(i)}\,=g(f_{\bm{\theta}}, \textit{\textbf{x}}_2^{(i)}\,),\textit{\textbf{x}}_2^{(i)}\in \textit{\textbf{X}}^{'}_2\}$
            \State Calculate procedural fairness loss $\mathcal{L}_{GPF}$ according to Equation (\ref{eq:l1-gpf})
            \State \raggedright Calculate the total loss $\mathcal{L}=\mathcal{L}_{CE}+\alpha\times \mathcal{L}_{GPF}$
            \State Update model parameters with $\bigtriangledown_{\bm{\theta}} \mathcal{L}$ using Adam
        \EndWhile 
        \State \Return the trained ML model $f_{\bm{\theta}}$
	\end{algorithmic} 
\end{algorithm}

First, the training set $\textit{\textbf{X}}$ is divided into $\textit{\textbf{X}}_{1}$ and $\textit{\textbf{X}}_{2}$ according to the sensitive attribute (lines 1-2). Then, for each data point, find the data point that is different from its sensitive attribute value but the most similar to it to get the set to be explained (lines 3-11). At this point, $\textit{\textbf{X}}^{'}_{1}$ and $\textit{\textbf{X}}^{'}_{2}$ contain all the training data and their most similar data points, where data points $\textbf{\textit{x}}_1^{(i)}\in \textit{\textbf{X}}^{'}_{1}$ and $\textbf{\textit{x}}_2^{(i)}\in \textit{\textbf{X}}^{'}_{2}$ corresponding to the subscripts are a pair of most similar data points. Subsequently, we start iterative training (lines 12-19). We first compute the cross-entropy loss $\mathcal{L}_{CE}$ (line 13) and the procedural fairness loss $\mathcal{L}_{GPF}$ of the current model (lines 14-16). Then, we obtain the total loss $\mathcal{L}$ (line 17) and update the model parameters (line 18).  At the end of the iteration, the trained ML model $f_{\bm{\theta}}$ is returned (line 20).

\subsection{Validation of the Proposed Method}
In the following subsection, we validate the effectiveness of above proposed method for achieving procedural fairness through comprehensive experiments. Firstly, we delineate the relevant experimental setting (including datasets and parameter setting) utilized both in this subsection and in Section \ref{sec:relationship}. Then, the baseline and evaluation metrics used in this subsection are introduced. Finally, the experimental results are presented and analyzed to validate the proposed method.

\subsubsection{Experimental Setup}
\label{sec:setup1}
\paragraph{Dataset}
\label{sec:dataset}
In this study, we conducted experiments on seven datasets, consisting of six widely-used real-world datasets commonly employed for fair ML~\citep{le2022survey}, and one synthetic dataset. The six real-world datasets include \textit{Adult}~\citep{Dua:2019}, \textit{Bank}~\citep{Dua:2019}, \textit{COMPAS}~\citep{angwin2016machine}, \textit{Default}~\citep{Dua:2019}, \textit{KDD}~\citep{Dua:2019}, and \textit{LSAT}~\citep{wightman1998lsac}. It's worth noting that for the three datasets that are inherently fairer (i.e., \textit{Bank}, \textit{Default}, and \textit{KDD}), with dataset \textit{DP} values less than 0.10, we randomly resampled the data points labeled “1" in the advantaged group until their dataset's \textit{DP} value exceeded 0.10. These resampled datasets were referred to as \textit{Unfair Bank}, \textit{Unfair Default}, and \textit{Unfair KDD}.

\begin{figure}
    \centering
    \includegraphics[width=0.5\linewidth]{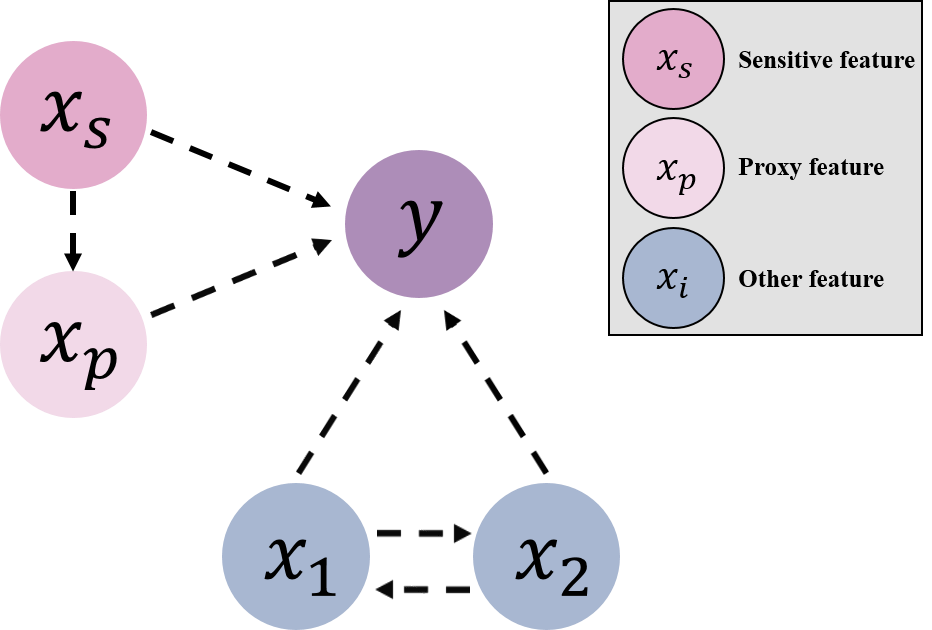}
    \caption{Relationship graph for the synthetic dataset.}
    \label{fig:synthetic_data}
\end{figure}

For the synthetic dataset, we referred to the data generation scheme of~\cite{zafar2019fairness} and~\cite{jones2020metrics}, resulting in a dataset of \num[group-separator={,}]{20000} data points. It includes two non-sensitive features $x_1$ and $x_2$, a sensitive feature $x_s$, a proxy feature $x_{p}$ for the sensitive feature $x_s$, and a labeling class $y$. Specifically, we first randomly generated \num[group-separator={,}]{20000} binary class labels $y$. Then, we drew samples from two different Gaussian distributions, associating features $x_1$ and $x_2$ with each label: $p([x_1, x_2]|y=1)=\mathcal{N}([2; 2], [5, 1; 1, 5])$, $p([x_1, x_2]|y=0)=\mathcal{N}([-2; -2], [10, 1; 1, 3])$. Additionally, the sensitive attribute $x_s$ was sampled from two different Bernoulli distributions: $p(x_s|y=1)=Bernoulli(p)$, $p(x_s|y=0)=Bernoulli(1-p)$. And the proxy feature $x_{p}\sim \mathcal{N} (x_s, 0.5)$. Obviously, the parameter $p$ controls the degree of bias in the dataset. In this section, we set $p$ to 0.65 and called the dataset \textit{Synthetic-0.65}. The relationship graph between features and the class label in the synthetic dataset is shown in Figure \ref{fig:synthetic_data}. Further details regarding the above seven datasets are summarized in Table \ref{tab:dataset}. We consistently preprocessed each dataset: label encoding of the categorical features, followed by normalization of all features with Z-score. In addition, each dataset was randomly divided into training and test sets in a ratio of 4:1.

\begin{table}[htbp]
    \centering
    \begin{tabular}{ccccccc}
        \toprule
            Dataset & $|\bm{\mathcal{D}}|$ & $|\textit{\textbf{X}}|$ & DP & $S$ & Advantaged & Disadvantaged \\
        \midrule
        Adult & \num[group-separator={,}]{48842} & 14 &  0.195 & Sex & Male & Female \\
        Bank & \num[group-separator={,}]{45211} & 16 & 0.048 & Marital & Single & Married \\
        COMAPS & \num[group-separator={,}]{6172} & 7 & 0.132 & Race & Others & Afi.-Am. \\
        Default & \num[group-separator={,}]{30000} & 23 & 0.034 & Sex & Male & Female \\
        KDD & \num[group-separator={,}]{284556} & 36 & 0.076 & Sex & Male & Female \\
        LSAT & \num[group-separator={,}]{20798} & 11 & 0.198 & Race & White & Non-White \\
        Synthetic-0.65 & \num[group-separator={,}]{20000} & 4 & 0.295 & $x_s$ & 1 & 0 \\
        \bottomrule
    \end{tabular}
    \caption{Seven datasets used in this paper. “$|\bm{\mathcal{D}}|$", “$|\textit{\textbf{X}}|$", “DP", “$S$", “Advantaged", and “Disadvantaged" denote the number of data points, the number of features, the \textit{DP} value of the dataset, the sensitive attribute under consideration, and the corresponding advantaged and disadvantaged groups, respectively.}
    \label{tab:dataset}
\end{table}

\paragraph{Parameter Setting}

In our experiments, consistent with~\cite{wang2024procedural}, a two-layer multi-layer perceptron (MLP) was trained with the ReLU activation function and Adam optimizer on each dataset. The total loss was composed of the binary cross-entropy (BCE) loss $\mathcal{L}_{CE}$ plus the procedural fairness loss $\mathcal{L}_{GPF}$, weighed by the parameter $\alpha$. Each model was fully connected to the hidden layer with 32 nodes, except for the \textit{KDD} datasets, which had a high number of features and we set the hidden layer with 64 nodes. The number of iterations and the learning rate were set to 300 and 0.01, respectively~\citep{wang2024procedural}. For the parameter $n$ (the number of pairs of similar data points) in the \textit{GPF\textsubscript{FAE}} metric, we similarly aligned with~\cite{wang2024procedural} and set it to 100. As for the hyperparameter $\alpha$, which weighs the BCE loss $\mathcal{L}_{CE}$ and the procedural fairness loss $\mathcal{L}_{GPF}$ in our method, we did not specifically tune the parameter on each dataset but set it uniformly to 0.5. For the distributive fairness metric \textit{DP}, we used the implementation in the open-source Python algorithmic fairness toolkit AI Fairness 360~\citep{bellamy2018ai}. All the experiments were performed with 10 independent runs by using different random number seeds.

\subsubsection{Baseline and Evaluation Metrics}

The baseline for our comparison was the MLP model trained using only BCE as the loss function (model architecture and other parameters remain unchanged), referred to as \textit{MLP\textsubscript{BCE}} in the following. The metrics evaluated include model accuracy (ACC), \textit{GPF\textsubscript{FAE}}, and \textit{DP}. That is, we compared the change in procedural and distributive fairness of the ML model before and after procedural fairness was taken into account, and the effect on model accuracy.

It is worth mentioning that the two methods proposed by~\cite{wang2024procedural} do not entail the development of procedurally-fair ML models; rather, they focus on enhancing the procedural fairness of an already trained ML model. Consequently, these methods are not suitable as baseline comparisons. Another potential baseline is the procedurally-fair ML model trained based on feature selection proposed by~\cite{grgic2018beyond}. However, their feature selection is based on the fairness of features assessed by human users, and they have only been studied and analyzed on two datasets, making it difficult to extend their method to other datasets for comparative experiments, and are therefore not compared. However, this does not preclude an assessment of the effectiveness of our approach, as a comparison with \textit{MLP\textsubscript{BCE}} methods is sufficient to know whether our approach improves the procedural fairness of the ML model and its impact on distributive fairness and model accuracy.

\subsubsection{Experimental Results}

Table~\ref{tab:method_compare} shows the results of comparing our method with \textit{MLP\textsubscript{BCE}} on the seven datasets for the ACC, \textit{GPF\textsubscript{FAE}}, and \textit{DP} metrics, from which we can obtain the following three conclusions:

\begin{enumerate}
    \item[(1)] \textbf{Our method significantly improves procedural fairness.} In particular, on datasets other than \textit{COMPAS} and \textit{LSAT}, our method improves the \textit{GPF\textsubscript{FAE}} metric from close to 0.0 to near or at 1.0. Even on the \textit{COMPAS} and \textit{LSAT} datasets, where \textit{MLP\textsubscript{BCE}} models initially have relatively higher \textit{GPF\textsubscript{FAE}} metric values, our method still improves the \textit{GPF\textsubscript{FAE}} metric values significantly.
    \item[(2)] \textbf{Our method has minimal impact on model performance.} The average loss in model accuracy on the seven datasets is 0.9$\%$. This indicates that our approach effectively enhances fairness without compromising model performance significantly.
    \item[(3)] \textbf{Distributive fairness also benefits substantially from our approach.} The distributive fairness metric \textit{DP} also improves considerably in the vast majority of datasets. Especially on the \textit{Unfair Bank}, \textit{Unfair Default}, \textit{Unfair KDD}, and \textit{LSAT} datasets, after optimizing the \textit{GPF\textsubscript{FAE}} metric, the \textit{DP} metric value has been reduced to a smaller range. This suggests that improving procedural fairness also contributes to improving distributive fairness.
\end{enumerate}

\begin{table}[t]
    \centering
    \begin{tabular}{c|ccc|ccc}
    \toprule
    \multirow{2}{*}{Dataset} & \multicolumn{3}{c|}{\textit{MLP\textsubscript{BCE}}} & \multicolumn{3}{c}{Our Method} \\
    \cmidrule{2-7}
    & ACC $\uparrow$ & $GPF_{FAE} \uparrow$ & \textit{DP} $\downarrow$ & ACC $\uparrow$ & $GPF_{FAE} \uparrow$ & \textit{DP} $\downarrow$ \\
    \midrule 
    Adult & \textbf{85.1$\%$} & 0.013 & 0.180 & 84.0$\%$ & \textbf{0.747} & \textbf{0.154} \\
    Unfair Bank & \textbf{86.2$\%$} & 0.002 & 0.150 & 83.0$\%$ & \textbf{0.993} & \textbf{0.069} \\
    COMPAS & 68.1$\%$ & 0.619 & \textbf{0.239} & \textbf{68.3$\%$} & \textbf{0.982} & 0.261 \\
    Unfair Default & \textbf{79.7$\%$} & 0.000 & 0.126 & 79.5$\%$ & \textbf{0.938} & \textbf{0.058} \\
    Unfair KDD & \textbf{94.0$\%$} & 0.000 & 0.104 & 92.8$\%$ & \textbf{0.952} & \textbf{0.026} \\
    LSAT & \textbf{89.9$\%$} & 0.422 & 0.201 & 89.5$\%$ & \textbf{0.808} & \textbf{0.063} \\
    Synthetic-0.65 & \textbf{88.6$\%$} & 0.049 & 0.336 & 87.2$\%$ & \textbf{1.000} & \textbf{0.222} \\
    \midrule
    \midrule
    Mean & \textbf{84.4$\%$} & 0.158 & 0.191 & 83.5$\%$ & \textbf{0.917} & \textbf{0.122} \\
    \bottomrule 
    \end{tabular}
    \caption{The evaluation results of two methods on ACC, \textit{GPF\textsubscript{FAE}}, and \textit{DP} metrics on each dataset. The better results of the two methods are bolded. $\uparrow$ means the larger the metric the better, and $\downarrow$ means the opposite. The average results for the seven datasets are shown at the bottom.}
    \label{tab:method_compare}
\end{table}

\begin{figure}[t]
    \centering
    \includegraphics[width=0.6\linewidth]{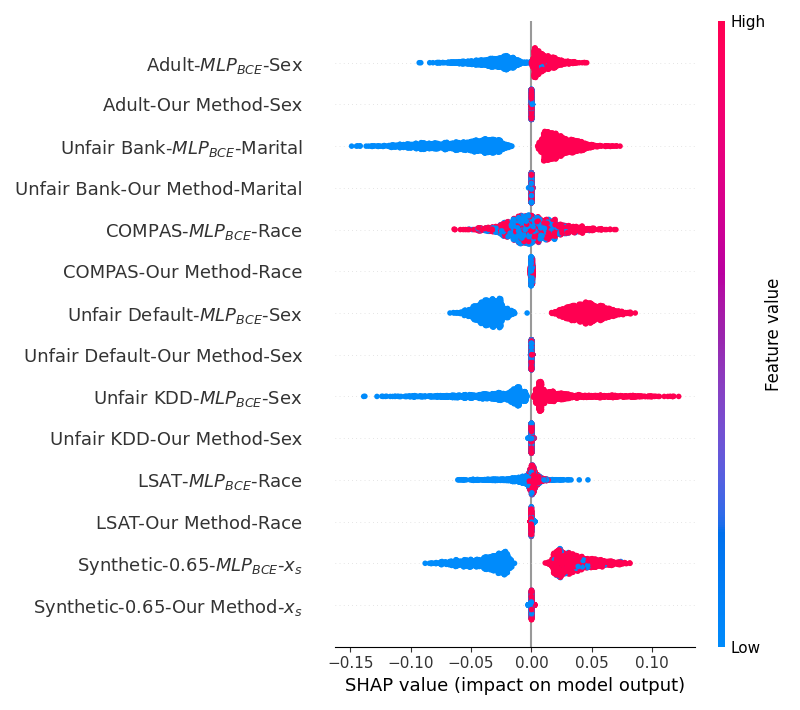}
    \caption{Comparing the \textit{SHAP} value of the sensitive attribute of the ML model obtained by the two methods on each dataset. Each red and blue dot represents a data point for an advantaged and disadvantaged group, respectively. The horizontal coordinate values are the results of the \textit{SHAP} method's explanation of the sensitive attribute of the corresponding data point. Larger absolute \textit{SHAP} values represent a greater impact on the decision, while a positive/negative sign indicates a positive/negative benefit to the decision. By employing our approach, the influence of sensitive attributes on the decision-making process is diminished to nearly zero, ensuring procedural fairness.}
    \label{fig:mlp_vs_pf_explain}
\end{figure}

We also visualize \textit{SHAP}~\citep{lundberg2017unified} explanations obtained by the two methods on the sensitive attribute to examine if and how the sensitive attribute has an impact on the decision process in both ML models. This is because the effect of the sensitive attribute on the decision-making process can also reflect the procedural fairness of the ML model side by side. Consistent with the data points used to evaluate the \textit{GPF\textsubscript{FAE}} metric, we visualize the \textit{SHAP} value of the sensitive attribute for set $\textit{\textbf{X}}^{'}=\textit{\textbf{X}}^{'}_1\cup \textit{\textbf{X}}^{'}_2$. That is, the distribution of explanations on the sensitive attribute for $n$ pairs of similar data points with a different sensitive attribute was visualized, as shown in Figure \ref{fig:mlp_vs_pf_explain}. Each row within the visualization corresponds to the explanation results of a particular method concerning the sensitive attribute within a dataset. For example, the first row, labeled “Adult-\textit{MLP\textsubscript{BCE}}-Sex", illustrates the visualization results of the \textit{MLP\textsubscript{BCE}} model concerning the “Sex" sensitive attribute within the \textit{Adult} dataset. Each red and blue dot represents a data point in $\textit{\textbf{X}}^{'}$ belonging to the advantaged and disadvantaged groups, respectively. The value corresponding to the horizontal coordinate of that data point is the result of the \textit{SHAP} method's interpretation of that data point on the sensitive attribute. It represents the impact (importance score) of that data point's sensitive attribute on its decision. Among them, the larger absolute \textit{SHAP} value represents its greater impact on decision-making, while the positive/negative sign represents its positive/negative benefit on decision-making.

As can be seen from Figure \ref{fig:mlp_vs_pf_explain}, for the \textit{MLP\textsubscript{BCE}} model, on datasets other than \textit{COMPAS} and \textit{LSAT}, it is evident that the advantaged group experiences positive benefits for decision-making, while the disadvantaged group incurs negative impacts. This indicates that there is a bias in favor of the advantaged group within the ML model's decision-making process. Conversely, on the \textit{COMPAS} and \textit{LSAT} datasets, there is no significant bias favoring either group. This observed phenomenon aligns with the performance of the \textit{GPF\textsubscript{FAE}} metric in Table \ref{tab:method_compare}. In contrast, after optimizing the \textit{GPF\textsubscript{FAE}} metric by our method, it can be seen that the sensitive attribute no longer affects the decision-making process on all datasets. Consequently, this is another confirmation of the effectiveness of our approach in improving the procedural fairness of the ML model.

In addition, although the method of obtaining FAE explanations was replaced from \textit{SHAP}~\citep{lundberg2017unified} to a more efficient gradient-based method, \textit{Grad}~\citep{simonyan2014deep}, during the training process, there is obviously a certain computational cost associated with the need to compute the explanation results of all the training data in each epoch. Thus, we compared the running time of our method with that of directly training an \textit{MLP\textsubscript{BCE}} model, and the results are shown in Table \ref{tab:time_compare}. The computing environment was a Linux server equipped with an AMD Ryzen Threadripper PRO 3995WX 64-core processor with 512GB RAM.

\begin{table}[htbp]
    \centering
    \begin{tabular}{l|c|ccc}
    \toprule
    Dataset & \textit{MLP\textsubscript{BCE}} & Our Method\\
    \midrule
    Adult & 4.26 & 8.62 \\
    Unfair Bank & 4.61 & 8.76 \\
    COMPAS & 1.13 & 2.01 \\
    Unfair Default & 3.74 & 7.56 \\
    Unfair KDD & 11.25 & 24.83 \\
    LSAT & 2.44 & 4.76 \\
    Synthetic-0.65 & 1.72 & 3.69 \\
    \midrule
    \midrule
    Mean & 4.16 & 8.60 \\
    \bottomrule 
    \end{tabular}
    \caption{Comparison of the average running times (in seconds) of training models on each dataset. The average results for the seven datasets are shown at the bottom.}
    \label{tab:time_compare}
\end{table}

Table \ref{tab:time_compare} clearly demonstrates that our method does introduce a moderate computational cost, with an average training time of approximately 8.6 seconds. This is due to the additional computation of an $\mathcal{L}_{GPF}$ term in its loss function relative to \textit{MLP\textsubscript{BCE}}, which takes about twice as long to train as \textit{MLP\textsubscript{BCE}}, but is still a relatively quick process.

Overall, in this section, we present the first method to achieve procedural fairness during training. It's a straightforward but effective method that takes into account the procedural fairness metric \textit{GPF\textsubscript{FAE}} in the form of a regularisation term during the training process. Experiments on one synthetic dataset and six real-world datasets show that our approach significantly improves the ML model's procedural fairness, and the distributive fairness is also significantly improved, while the impact on the ML model's accuracy is only 0.9$\%$ on average with a modest time cost.

\section{Relationship between Procedural and Distributive Fairness}
\label{sec:relationship}

In this section, we analyze the relationship between the ML model's procedural fairness and distributive fairness with the help of the approach presented in Section \ref{sec:method}. We demonstrate that, in essence, the distributive unfairness observed in ML models is merely symptomatic, whose roots originate in both the dataset and the decision-making processes of the ML models. Therefore, our objective is to analyze the impact on the distributive fairness of the ML model in the presence or absence of bias in the dataset and in the decision-making process of the ML model. 

For both distributive fairness and procedural fairness in ML, we now have corresponding metrics and optimization methods. It is important to clarify the distinctions between these two dimensions. Therefore, this section formulates and addresses two research questions, comprising four and two sub-questions, respectively, as described below:

\begin{enumerate} [leftmargin=27pt]
    \item[\textbf{\textit{RQ1}}] What are the influences of inherent dataset bias and the ML model's procedural fairness on its distributive fairness?
        \begin{enumerate}
            \item[\textit{RQ1.1}] Does the ML model exhibit distributive fairness when the dataset is unbiased and the ML model is procedurally-fair?
            \item[\textit{RQ1.2}] How does the bias of the dataset affect the ML model's distributive fairness?
            \item[\textit{RQ1.3}] How does procedural unfairness of the ML model affect its distributive fairness?
            \item[\textit{RQ1.4}] When the dataset is biased and the ML model is procedurally-unfair, what is the impact on its distributive fairness if they exhibit the same or opposite biases?
        \end{enumerate}
    \item[\textbf{\textit{RQ2}}] What are the differences between optimizing procedural fairness metrics and distributive fairness metrics in ML?
        \begin{enumerate}
            \item[\textit{RQ2.1}] How does optimization of procedural fairness metrics affect the fairness of the ML model?
            \item[\textit{RQ2.2}] How does optimization of distributive fairness metrics affect the fairness of the ML model?
        \end{enumerate}
\end{enumerate}

\subsection{Experimental Setup}

To address the proposed research questions, we need to complement the experimental setup in Section \ref{sec:setup1}. To answer \textit{RQ1}, we need biased and unbiased datasets as well as procedurally-fair and -unfair ML models, as described below.

\begin{enumerate}
    \item[(1)] \textbf{Unbiased datasets:} For the synthetic dataset, which has ground truth, we can easily ensure that it is unbiased by setting the parameter $p$ to 0.5 and calling the dataset \textit{Synthetic-0.5}. However, for the six real-world datasets, where ground truth information is unavailable, we preprocessed the datasets using LFR~\citep{zemel2013learning} to remove the information related to the sensitive attribute from the datasets. The transformed datasets are considered as unbiased datasets and are referred to as \textit{LFR-dataset names} (e.g., \textit{LFR-Adult}).
    \item[(2)] \textbf{Biased datasets:} Consistent with those in Section \ref{sec:dataset}, we considered \textit{Adult}, \textit{Unfair Bank}, \textit{COMPAS}, \textit{Unfair Default}, \textit{Unfair KDD}, \textit{LSAT}, and \textit{Synthetic-0.65} as biased datasets.
    \item[(3)] \textbf{Procedurally-fair ML models:} We utilized the method proposed in Section \ref{sec:method} to ensure that the decision-making process of the obtained ML model is fair and considered the obtained ML models as procedurally-fair ML models.
    \item[(4)] \textbf{Procedurally-unfair ML models:} We also utilized the method proposed in Section \ref{sec:method}. However, we inversely optimized the procedural fairness loss $\mathcal{L}_{GPF}$ (with the parameter $\alpha$ slightly less than 0). This is to ensure that the decision-making process of the obtained ML models is unfair and to consider the ML model obtained by back-optimizing the procedural fairness loss $\mathcal{L}_{GPF}$ as a procedurally-unfair ML model.
\end{enumerate}

In this section, the parameter $\alpha$ used to weigh $\mathcal{L}_{CE}$ and $\mathcal{L}_{GPF}$ in the loss function is case-specific and will be specified later. To address \textit{RQ2}, in addition to optimizing procedural fairness metric \textit{GPF\textsubscript{FAE}} using the method presented in Section \ref{sec:method}, we also used the following three in-process methods to consider the distributive fairness metric \textit{DP}: regularization-based~\citep{kamishima2012fairness}, constraint-based~\citep{agarwal2018reductions}, and adversarial learning-based~\citep{zhang2018mitigating}. This allows us to explore the difference between considering procedural fairness metrics and distributive fairness metrics in ML.

\subsection{RQ1 What Are the Influences of Inherent Dataset Bias and the ML Model's Procedural Fairness on its Distributive Fairness?}

In this section, we conduct comprehensive experiments to analyze whether and how dataset bias and the ML model's procedural fairness affect its distributive fairness. Our aim is twofold: to clarify their relationship and to assist in tracing the root causes of distributive unfairness.

\textit{RQ1.1 Does the ML Model Exhibit Distributive Fairness When the Dataset Is Unbiased and the ML Model Is Procedurally-Fair?}

We used \textbf{unbiased datasets} to train \textbf{procedurally-fair ML models} to assess the ML model's distributive fairness in this scenario. Since the dataset is unbiased, we can easily obtain the procedurally-fair ML models by setting the parameter $\alpha$ to 0.1 uniformly across the datasets. The performance of the ML model's procedural fairness metric \textit{GPF\textsubscript{FAE}} and distributive fairness metric \textit{DP} in this case is shown in Table \ref{tab:fairData-fairModel}.

\begin{table}[htbp]
    \centering
    \begin{tabular}{l|c|c}
    \toprule
    Dataset  & \textit{GPF\textsubscript{FAE}} $\uparrow$ & \textit{DP} $\downarrow$ \\
    \midrule
    \multirow{1}{*}{LFR-Adult} & 1.0 & 0.018 \\
    \multirow{1}{*}{LFR-Bank} & 1.0 & 0.003 \\
    \multirow{1}{*}{LFR-COMPAS} & 1.0 & 0.053 \\
    \multirow{1}{*}{LFR-Default} & 1.0 & 0.007 \\
    \multirow{1}{*}{LFR-KDD} & 1.0 & 0.006 \\
    \multirow{1}{*}{LFR-LSAT} & 0.991 & 0.002 \\
    \multirow{1}{*}{Synthetic-0.5} & 1.0 & 0.012 \\
    \bottomrule 
    \end{tabular}
    \caption{Performance of the \textit{GPF\textsubscript{FAE}} and \textit{DP} metrics when the dataset is unbiased and the ML model is procedurally-fair. $\uparrow$ means the larger the metric the better, and $\downarrow$ means the opposite.}
    \label{tab:fairData-fairModel}
\end{table}

From Table \ref{tab:fairData-fairModel}, we can see that the \textit{GPF\textsubscript{FAE}} metric is at or close to 1.0, which indicates that the ML model we have obtained is indeed procedurally-fair. In contrast, the \textit{DP} metric is very small on all datasets, i.e., when the dataset is unbiased and the ML model is procedurally-fair, it is also distributively-fair.

\textit{RQ1.2 How Does the Bias of the Dataset Affect the ML Model’s Distributive Fairness?}

To examine the effect of dataset bias on the ML model's distributive fairness, we utilized \textbf{biased datasets} to train \textbf{procedurally-fair ML models}. The purpose is to ensure that the observed distributive unfairness comes entirely from biases inherent in the dataset. On the datasets other than the \textit{Adult} and \textit{LSAT}, we were consistent with Section \ref{sec:method} and set the parameter $\alpha$ to 0.50. While on the \textit{Adult} and \textit{LSAT} datasets, the parameter $\alpha$ was set to 3.0 and 2.0, respectively. This is because, as can be seen from Table \ref{tab:method_compare}, our method yields smaller \textit{GPF\textsubscript{FAE}} metrics on the \textit{Adult} and \textit{LSAT} datasets compared to the remaining datasets. To confidently attribute bias originating solely from the datasets, we set larger $\alpha$ values on these two datasets to ensure the ML model is procedurally-fair. In addition, we subjected the experimental results to the \textit{DP} metric obtained from the \textbf{procedurally-fair ML models} trained using the \textbf{unbiased datasets} (i.e., the corresponding results in Table \ref{tab:fairData-fairModel}) using the Wilcoxon rank-sum test with a 0.05 significance level. This is to observe whether dataset bias has a statistically significant impact on the \textit{DP} metric, and the results are shown in Table \ref{tab:unfairData-fairModel}.

\begin{table}[htbp]
		\centering
		\begin{tabular}{l|c|c}
		\toprule
        Dataset & \textit{GPF\textsubscript{FAE}} $\uparrow$ & \textit{DP} $\downarrow$ \\
        \midrule
        \multirow{1}{*}{Adult} & 0.887 & 0.055 - \\
        \multirow{1}{*}{Unfair Bank} & 0.993 & 0.069 - \\
        \multirow{1}{*}{COMPAS} & 0.982 & 0.261 - \\
        \multirow{1}{*}{Unfair Default} & 0.938 & 0.058 - \\
        \multirow{1}{*}{Unfair KDD} & 0.952 & 0.026 - \\
        \multirow{1}{*}{LSAT} & 0.884 & 0.004 - \\
        \multirow{1}{*}{Synthetic-0.65} & 1.0 & 0.222 - \\
        \bottomrule 
		\end{tabular}
        \caption{Performance of the \textit{GPF\textsubscript{FAE}} and \textit{DP} metrics when the dataset is biased and the ML model is procedurally-fair. $\uparrow$ means that the larger the metric the better, and $\downarrow$ means the opposite. “-" indicates a statistically significant distributive unfairness on a Wilcoxon rank-sum test with a 0.05 significance level against the corresponding \textit{DP} metric in Table \ref{tab:fairData-fairModel}.}
		\label{tab:unfairData-fairModel}
	\end{table} 

As can be seen in Table \ref{tab:unfairData-fairModel}, the decision-making process of the ML model is very fair, and the effect of dataset bias on the ML model's distributive fairness is statistically significant. However, the extent of this impact varies significantly numerically across datasets. Notably, the effect is pronounced in the \textit{COMPAS} and \textit{Synthetic-0.65} datasets, while the \textit{DP} metric values do not exhibit substantial changes in the other five datasets.  Consequently, we conclude that dataset bias indeed indirectly affects the distributive fairness of ML models, albeit to a varying degree depending on the specific dataset, with an overall modest impact.

\begin{figure}[htpb]
    \centering
    \includegraphics[width=0.5\linewidth]{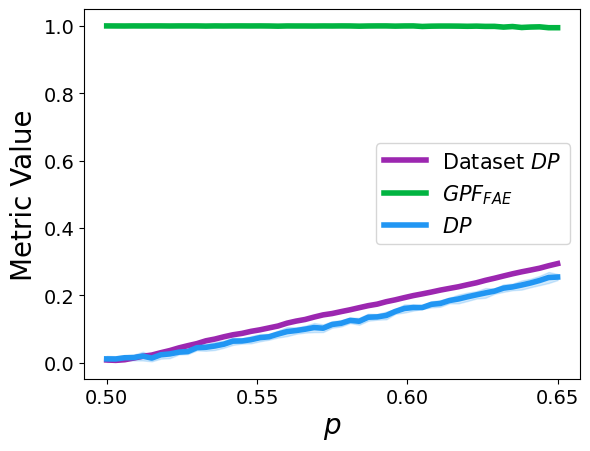}
    \caption{The trends of the dataset bias (Dataset \textit{DP}), distributive fairness (\textit{DP}), and procedural fairness (\textit{GPF\textsubscript{FAE}}) metrics with the increase of parameter $p$ which controls the degree of bias in the dataset.}
    \label{fig:relationship_data}
\end{figure}

However, as indicated in Table \ref{tab:unfairData-fairModel}, the \textit{GPF\textsubscript{FAE}} metric does not achieve a value of 1.0 across the six real-world datasets. Therefore, to further substantiate the conclusion that the bias of the datasets indeed affects the ML model's distributive fairness, we conducted further experiments focusing on the synthetic datasets. Specifically, we constructed synthetic datasets with various degrees of unfairness by controlling the parameter $p$, and trained \textbf{procedurally-fair ML models} by keeping the settings consistent with those above. Among them, the parameter $p$ was selected uniformly from 50 parameters between [0.5, 0.65]. Ultimately, the trends of each metric on the synthetic dataset with increasing bias in the dataset are shown in Figure \ref{fig:relationship_data}.

As can be seen in Figure \ref{fig:relationship_data}, the dataset gradually becomes unfair as the parameter $p$ increases, and the distributive fairness metric \textit{DP} steadily increases despite the fact that the ML model is always procedurally-fair. This further illustrates that dataset bias does affect the ML model's distributive fairness.

\textit{RQ1.3 How Does Procedural Unfairness of the ML Model Affect Its Distributive Fairness?}

In order to explore the impact of the ML model's procedural unfairness on its distributive fairness, we observe the distributive fairness of procedurally-unfair ML models that are trained on unbiased and biased datasets, respectively.

\paragraph{a) Distributive fairness of procedurally-unfair ML models trained on unbiased data:}

We used \textbf{unbiased datasets} to train \textbf{procedurally-unfair ML models} to ensure that the bias comes only from the ML model's decision process. However, on the six real-world datasets after preprocessing with LFR~\citep{zemel2013learning}, the ML models trained by backward optimizing the procedural fairness loss $\mathcal{L}_{GPF}$ are still procedurally-fair. This is because LFR~\citep{zemel2013learning} obfuscates the sensitive attribute information, making the data representations independent of the sensitive attribute. Consequently, ML models trained on such datasets, which do not contain group information, naturally no longer contain any group-related information and cannot be biased towards either group.

Therefore, we performed additional processing on the LFR-processed real-world datasets to reintroduce group information. For each real-world dataset, we performed two different treatments: 

\begin{enumerate}
\item[(i)] We reintroduced the true sensitive attribute (TSA) $s$ to the LFR-processed dataset, designating it as \textit{LFR-TSA-dataset name} (e.g., \textit{LFR-TSA-Adult}). The only drawback is that although the TSA is uncorrelated with other LFR-processed features, it is correlated with the class label, rendering the dataset not entirely unbiased.

\item[(ii)] We introduced a new fake sensitive attribute (FSA), $s_{fake}$, to the LFR-processed dataset. $s_{fake}$ was generated through Bernoulli sampling: $s_{fake}=Bernoulli (0.5)$, allocating each data point randomly to the advantaged and disadvantaged groups with equal probability. We then considered the fairness on the FSA and called the dataset \textit{LFR-FSA-dataset name} (e.g., \textit{LFR-FSA-Adult}). Notably, the FSA is generated randomly and is entirely independent of other features and the class label, rendering it a dataset that not only contains information about the sensitive attribute but also is absolutely unbiased.
\end{enumerate}

\begin{figure}[t]
    \centering
    \subfigure[\textit{LFR-TSA} dataset]{
    \includegraphics[width=0.3\linewidth]{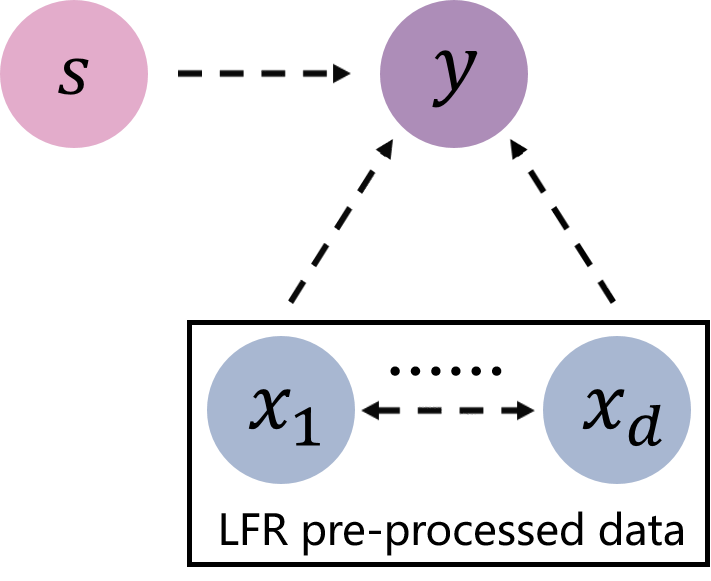}}
    \subfigure[\textit{LFR-FSA} Dataset]{
    \label{fig:LFR_data_b}
    \includegraphics[width=0.3\linewidth]{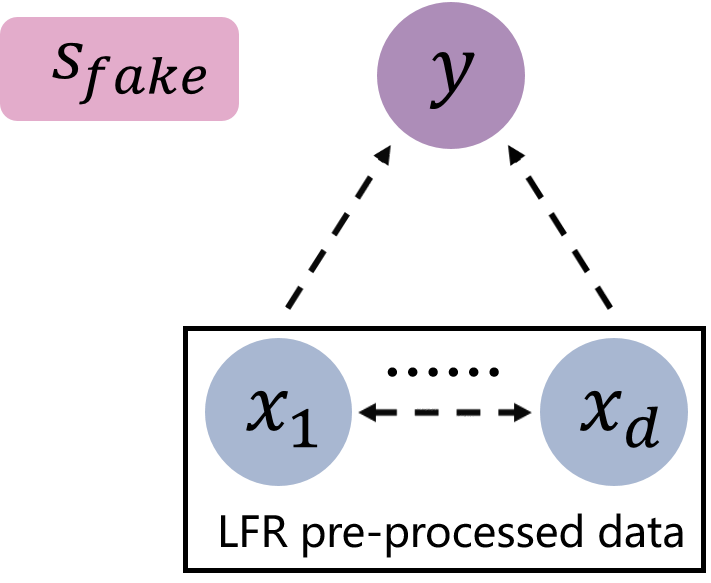}
    }
    \caption{Relationship graph for the \textit{LFR-TSA} and \textit{LFR-FSA} datasets.}
    \label{fig:LFR_data}
\end{figure}

The relationships between features and the class label resulting from these two treatments on the six real-world datasets are illustrated in Figure \ref{fig:LFR_data}. Different from the six real-world datasets, for the synthetic dataset, no additional processing is required as the ground truth is known. Unbiasedness is ensured by setting the parameter $p$ to 0.5 while retaining group information. Finally, the procedurally-unfair ML models were obtained by back-optimizing the procedural fairness loss $\mathcal{L}_{GPF}$ on the datasets described above. The parameter $\alpha$ was uniformly set to -0.02 and -0.10 on the \textit{LFR-TSA} and \textit{LFR-FSA} datasets, respectively. On the \textit{Synthetic-0.5} dataset, the $\alpha$ was set to -0.40. The experimental results are shown in Table \ref{tab:fairData_unfairModel}.

\begin{table}[t]
    \centering
    \begin{tabular}{l|c|c}
    \toprule
    Dataset & \textit{GPF\textsubscript{FAE}} $\uparrow$ & \textit{DP} $\downarrow$ \\
    \midrule
    LFR-TSA-Adult & 0.0 & 0.360 \\
    LFR-FSA-Adult & 0.013 & 0.375 \\
    \midrule
    LFR-TSA-Bank & 0.047 & 0.450 \\
    LFR-FSA-Bank & 0.003 & 0.284 \\
    \midrule
    LFR-TSA-COMPAS & 0.0 & 0.319 \\
    LFR-FSA-COMPAS & 0.008 & 0.328 \\
    \midrule
    LFR-TSA-Default & 0.010 & 0.204 \\
    LFR-FSA-Default & 0.0 & 0.466 \\
    \midrule
    LFR-TSA-KDD & 0.0 & 0.249 \\
    LFR-FSA-KDD & 0.0 & 0.206 \\
    \midrule
    LFR-TSA-LSAT & 0.003 & 0.439 \\
    LFR-FSA-LSAT & 0.0 & 0.438 \\
    \midrule
    \multirow{1}{*}{Synthetic-0.5}& 0.0 & 0.124 \\
    \bottomrule 
    \end{tabular}
    \caption{Performance of the \textit{GPF\textsubscript{FAE}} and \textit{DP} metrics when the dataset is unbiased and the ML model is procedurally-unfair. $\uparrow$ means the larger the metric the better, and $\downarrow$ means the opposite.}
    \label{tab:fairData_unfairModel}
\end{table}

Table \ref{tab:fairData_unfairModel} indicates that on the seven datasets, regardless of the way they are constructed, when the dataset is unbiased and the ML model is procedurally-unfair, the value of the distributive fairness metric \textit{DP} is very large. This implies that the ML model's procedural unfairness has a significant effect on its distributive fairness.

\paragraph{b) Distributive fairness of procedurally-unfair ML models trained on biased data:}

We also conducted experiments using \textbf{biased datasets} to train \textbf{procedurally-unfair ML models}, where the parameter $\alpha$ that weighs the $\mathcal{L}_{CE}$ and the $\mathcal{L}_{GPF}$ is uniformly set to -0.02. The results are then compared to the \textbf{procedurally-fair ML models} trained on the \textbf{biased datasets} (i.e., the corresponding results in Table \ref{tab:unfairData-fairModel}) using the Wilcoxon rank-sum test with a significance level of 0.05 to observe the effect of the ML model's procedural unfairness on its distributive fairness, and the results are shown in Table \ref{tab:unfairData-unfairModel}.

\begin{table}[htbp]
    \centering
    \begin{tabular}{l|c|c}
    \toprule
    Dataset & \textit{GPF\textsubscript{FAE}} $\uparrow$ & \textit{DP} $\downarrow$ \\
    \midrule
    \multirow{1}{*}{Adult} & 0.011 & 0.178 - \\
    \multirow{1}{*}{Unfair Bank} & 0.0 & 0.203 - \\
    \multirow{1}{*}{COMPAS} & 0.039 & 0.284 -  \\
    \multirow{1}{*}{Unfair Default} & 0.023  & 0.146 - \\
    \multirow{1}{*}{Unfair KDD} & 0.0 &  0.170 - \\
    \multirow{1}{*}{LSAT} & 0.0 & 0.255 - \\
    \multirow{1}{*}{Synthetic-0.65} & 0.014 & 0.342 - \\
    \bottomrule 
    \end{tabular}
    \caption{Performance of the \textit{GPF\textsubscript{FAE}} and \textit{DP} metrics when the dataset is biased and the ML model is procedurally-unfair. $\uparrow$ means the larger the metric the better, and $\downarrow$ means the opposite. “-" indicates a statistically significant distributive unfairness on a Wilcoxon rank-sum test with a 0.05 significance level against the corresponding \textit{DP} metric in Table \ref{tab:unfairData-fairModel}.}
    \label{tab:unfairData-unfairModel}
\end{table}

Table \ref{tab:unfairData-unfairModel} reveals a substantial increase in the \textit{DP} metric compared to Table \ref{tab:unfairData-fairModel}, following the decision-making process becomes unfair. This observation reinforces our earlier conclusion, highlighting the substantial influence of the ML model's procedural fairness on its distributive fairness. In addition, by examining Tables \ref{tab:fairData-fairModel}, \ref{tab:unfairData-fairModel}, and \ref{tab:unfairData-unfairModel} collectively, it becomes evident that the ML model's procedural fairness exerts a significantly greater influence on its distributive fairness than that of dataset bias. Moreover, the experimental results of \textit{RQ1.1-RQ1.3} confirm that the ML model's distributive unfairness stems precisely from the inherent bias of the dataset and the ML model's procedural unfairness.

\textit{RQ1.4 When the Dataset Is Biased and the ML Model Is Procedurally-Unfair, What Is the Impact on its Distributive Fairness If They Exhibit the Same or Opposite Biases?}

To answer this question, we need to obtain datasets and decision processes for ML models with the same and different biases. Obviously, \textbf{biased datasets} are in favor of the advantaged group. Consequently, our objective is to construct ML models in which the decision processes are biased towards the advantaged and disadvantaged groups, respectively. This allows us to observe the impact on distributive fairness when the two have the same and different biases. However, achieving this purpose on black-box ML models poses considerable challenges.

Therefore, referring to~\cite{wang2024procedural}, we first performed feature selection on the \textbf{biased dataset} by retaining only the fair features and the sensitive attribute, thus ensuring that the bias in the dataset comes only from the sensitive attribute. Specifically, on the \textit{Synthetic-0.65} dataset, we used features $x_1$, $x_2$, and the sensitive attribute $x_s$. Whereas on the real-world datasets, since there is no ground truth, features with a Pearson correlation coefficient of less than 0.10 with the sensitive attribute were considered fair features~\citep{wang2024procedural}. An exception was made for the \textit{COMPAS} and \textit{LSAT} datasets, given their heightened correlation between each feature and the sensitive attribute, where a threshold of 0.20 was applied~\citep{wang2024procedural}.

Following feature selection, we trained a logistic regression linear model using the resulting dataset. Subsequently, on the trained linear model, we manually controlled the weight assigned to the sensitive attribute, denoted as $w_s$. Obviously, when $w_s$ is 0.0, the sensitive attribute does not affect the linear model's decision-making process, so it is a procedurally-fair ML model. And $w_s$ greater/less than 0.0 means that the linear model favors the advantaged/disadvantaged group. And the magnitude of the absolute value of $w_s$ represents the degree of the ML model's procedural unfairness. 

\begin{figure}[htpb]
    \centering
    \subfigure[\textit{GPF\textsubscript{FAE}} metric]{
		\includegraphics[width=0.45\linewidth]{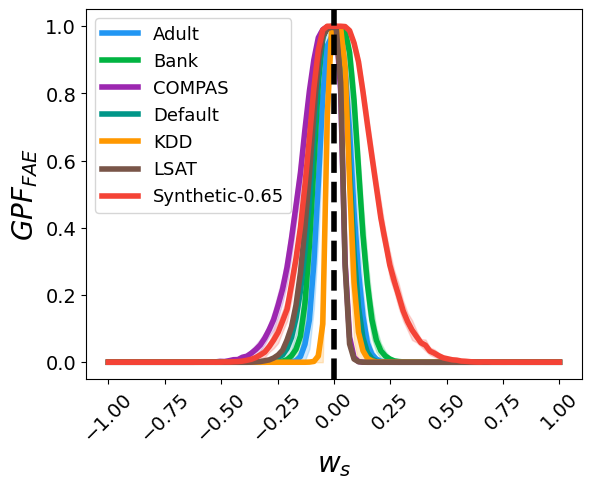}}
    \subfigure[\textit{DP} metric]{
		\includegraphics[width=0.45\linewidth]{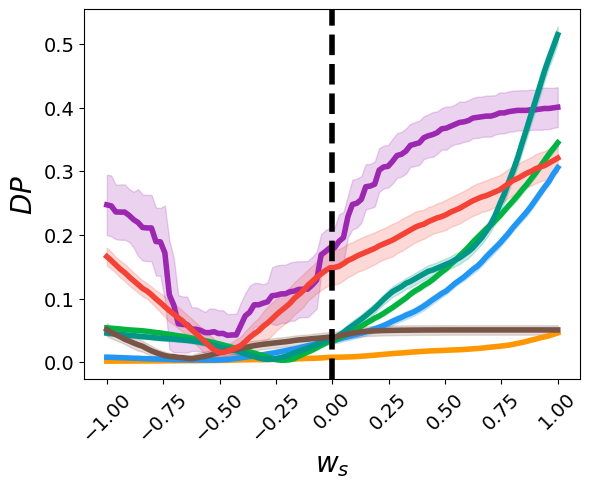}}
    \caption{The trends of \textit{GPF\textsubscript{FAE}} (procedural fairness) and \textit{DP} (distributive fairness) metrics with changes in $w_s$. A larger absolute value of $w_s$ means that the decision-making process of the ML model is more unfair, and a positive/negative sign indicates that the ML model is biased in favor of the advantaged/disadvantaged group. Figures show mean values over 10 random runs with a $1-\sigma$ error-bar. Here, $\sigma$ is the standard deviation.}
    \label{fig:dif_bias}
\end{figure}

Therefore, we first investigated the impact on the ML model's distributive fairness when the dataset favors the advantaged group and the ML model has various degrees of bias against the advantaged or disadvantaged group by controlling the value of the parameter $w_s$. The range of $w_s$ values was set to [-2, 2], [-2, 2], [-2, 2], [-1.5, 1.5], [-2, 2], [-2, 2], [-2, 2], [-5, 5], corresponding to different datasets. A total of 100 parameters were generated uniformly within each range to construct procedurally-unfair ML models with various degrees of bias. Finally, we normalized the parameter $w_s$ for each dataset to facilitate the presentation of the experimental results. The trends of  \textit{GPF\textsubscript{FAE}} and \textit{DP} metrics with $w_s$ on each dataset are shown in Figure \ref{fig:dif_bias}.

From Figure \ref{fig:dif_bias}, we can see that when $w_s$ is 0.0, the \textit{GPF\textsubscript{FAE}} metric reaches or is very close to 1.0, i.e., the ML model is procedurally-fair at this point. However, the distributive fairness metric \textit{DP} is not 0.0, which further validates our conclusion in \textit{RQ1.2} that the inherent bias of the dataset can lead to distributive unfairness of the ML model. And when $w_s$ is greater than 0.0, that is the dataset and the ML model's decision-making process exhibit the same bias, the \textit{GPF\textsubscript{FAE}} metric keeps decreasing and the \textit{DP} metric keeps increasing as $w_s$ increases. This means that when the two have the same bias, the bias is superimposed thereby exacerbating the ML model's distributive unfairness.

However, when the dataset and the ML model's decision-making process show opposite biases (i.e., $w_s$ is less than 0.0), the \textit{GPF\textsubscript{FAE}} metric still keeps decreasing as the absolute value of $w_s$ increases. However, the \textit{DP} metric initially decreases before subsequently increasing. This means that when the two have opposite biases, the biases of the two will cancel each other out, and even the phenomenon of equilibrium makes the ML model exhibit distributive fairness.

\begin{figure}[htpb]
    \centering
    \includegraphics[width=0.5\linewidth]{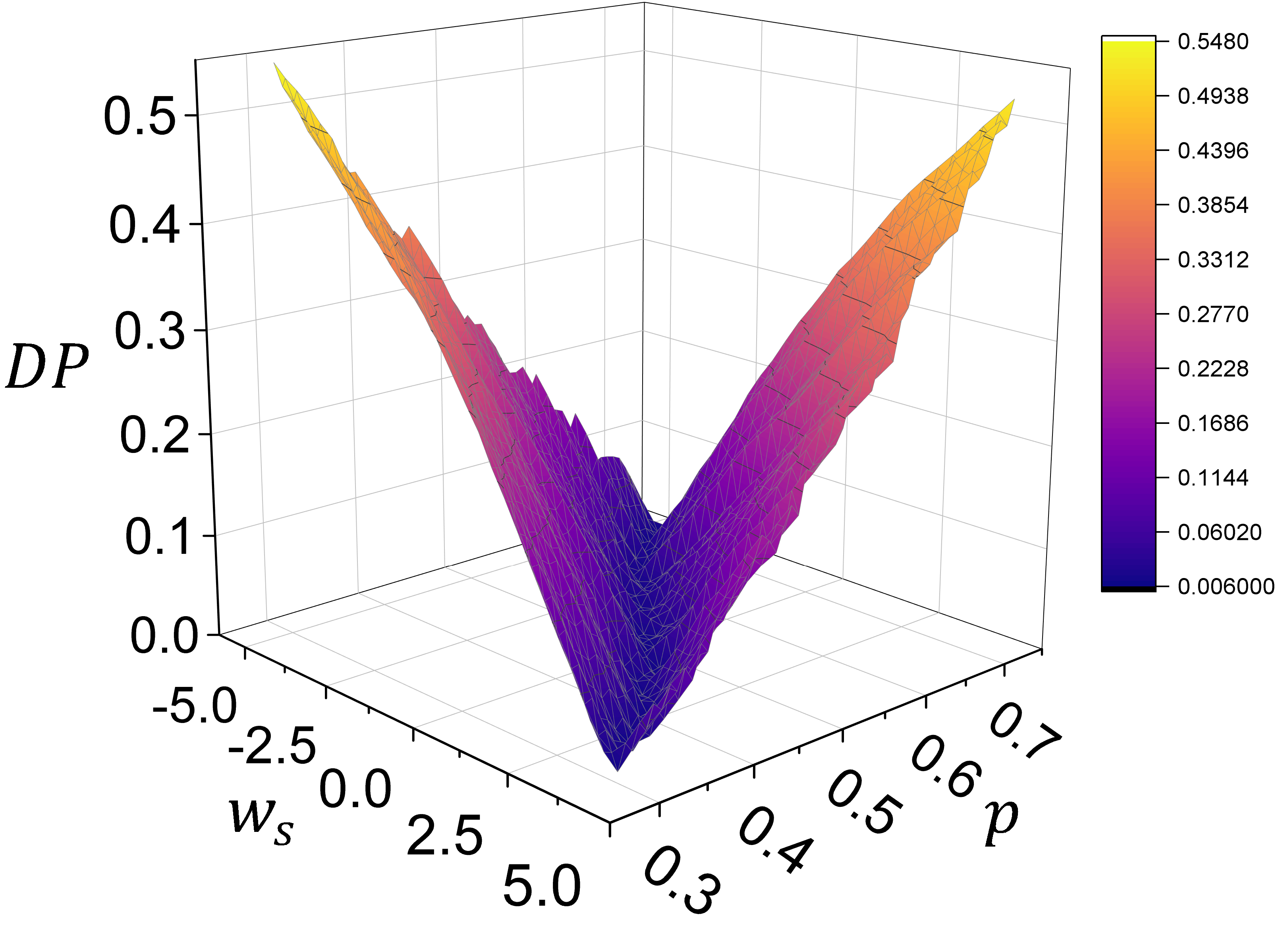}
    \caption{Trends in the \textit{DP} metric with changes in the parameters $p$ and $w_s$. When parameters $p$ and $w_s$ approach 0.5 and 0.0 respectively, it indicates fairness in the data and decision-making process of the ML model; whereas, parameters $p$ and $w_s$ being greater than/less than 0.5 and 0.0 respectively signify bias favoring advantaged/disadvantaged groups in both data and ML model decision-making processes.}
    \label{fig:dif_bias_3D}
\end{figure}

\begin{figure}[t]
    \centering
    \subfigure[$p=0.4$]{
    \label{fig:dif_bias_3D_intercept_a}
    \includegraphics[width=0.35\linewidth]{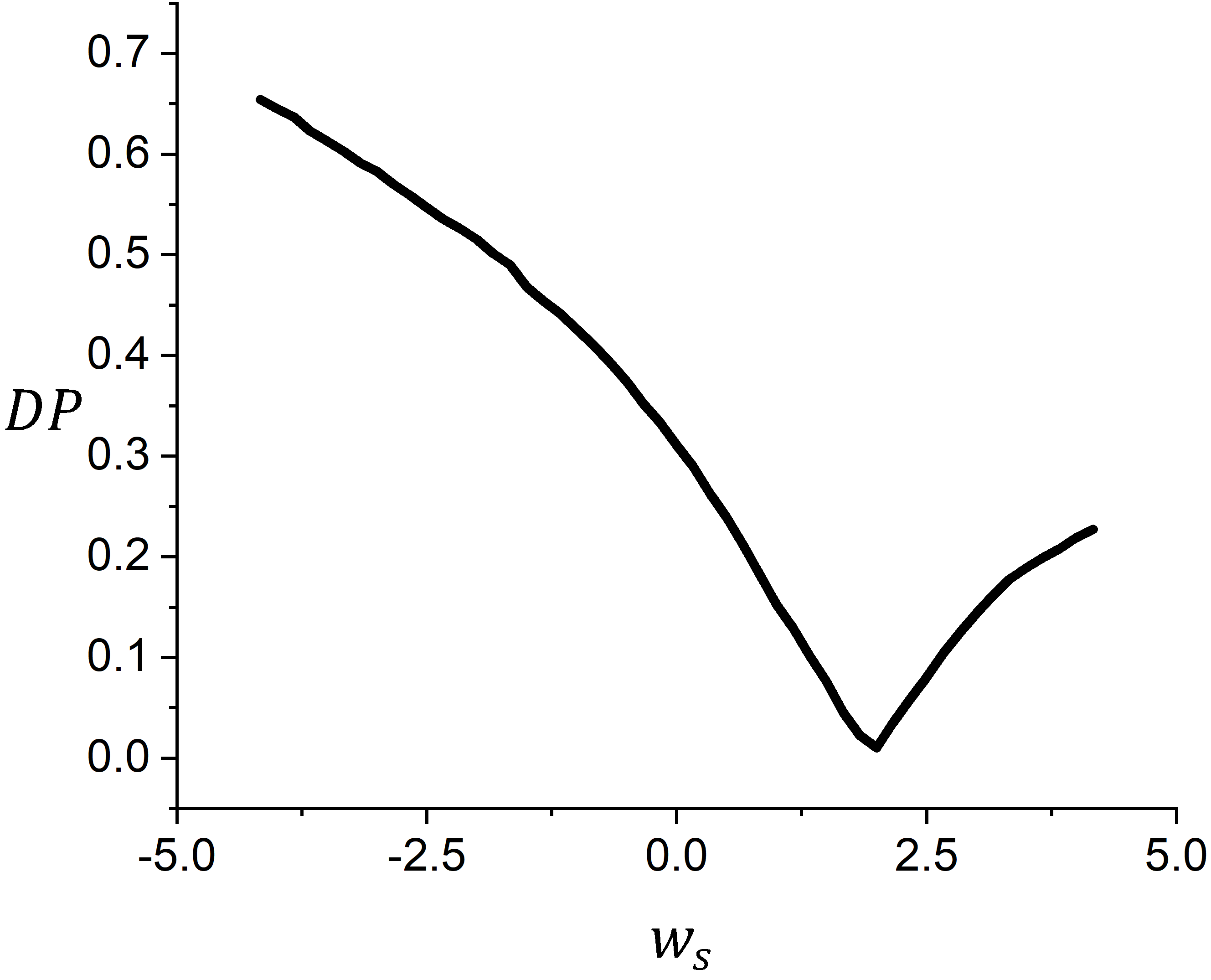}
    }
    \subfigure[$p=0.6$]{
    \label{fig:dif_bias_3D_intercept_b}
    \includegraphics[width=0.35\linewidth]{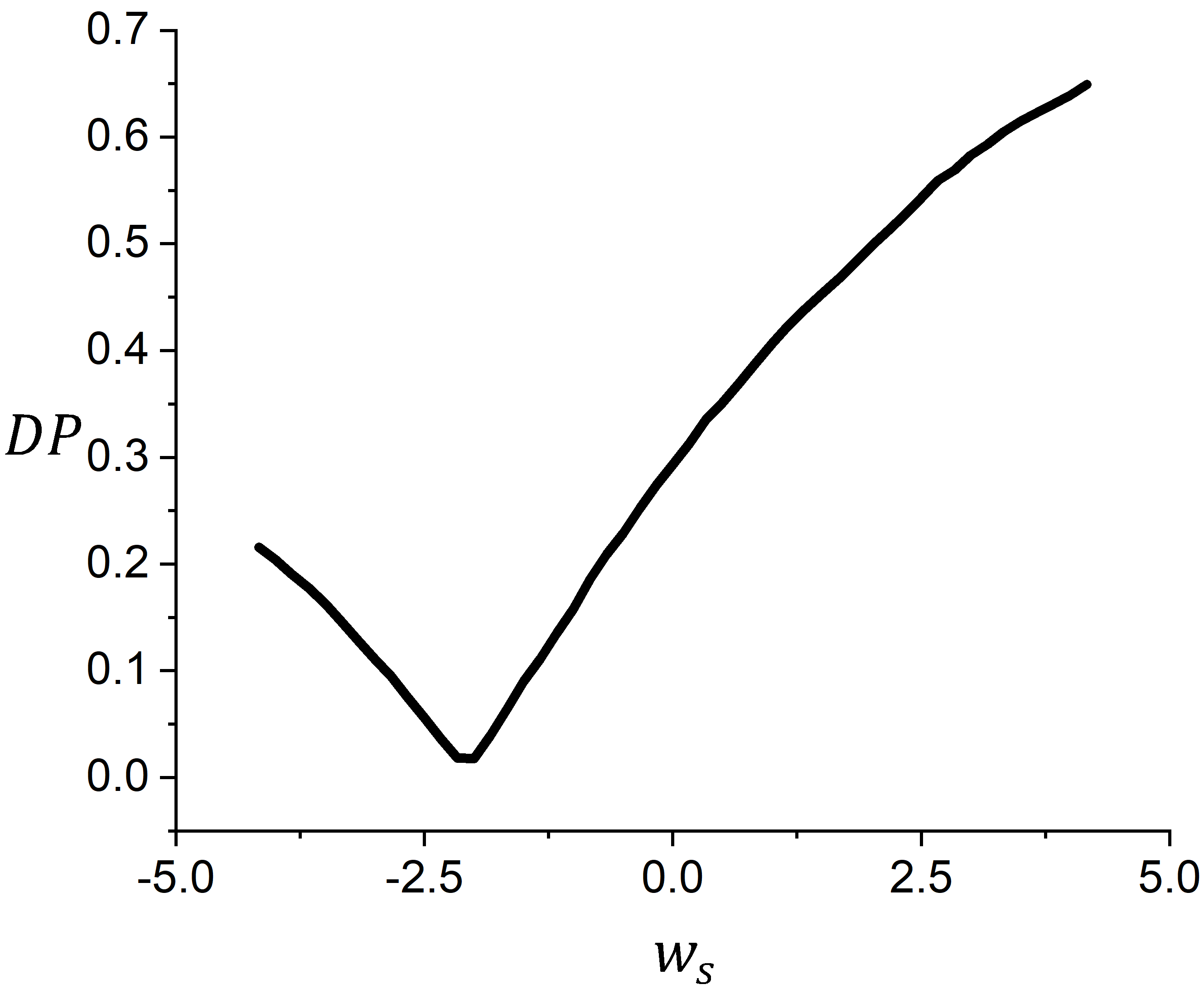}
    }
    \subfigure[$w_s=-1.0$]{
    \label{fig:dif_bias_3D_intercept_c}
    \includegraphics[width=0.35\linewidth]{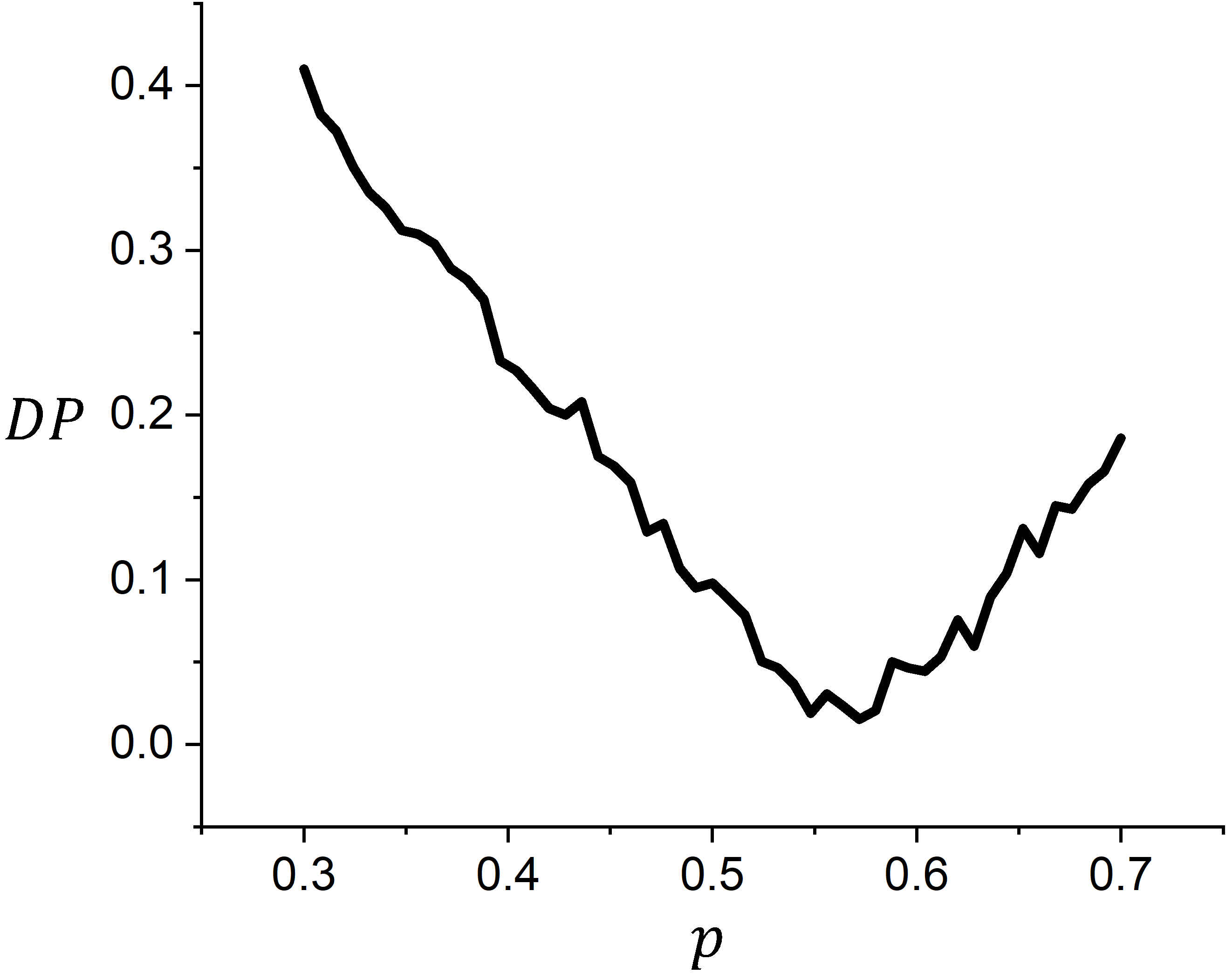}
    }
    \subfigure[$w_s=1.0$]{
    \label{fig:dif_bias_3D_intercept_d}
    \includegraphics[width=0.35\linewidth]{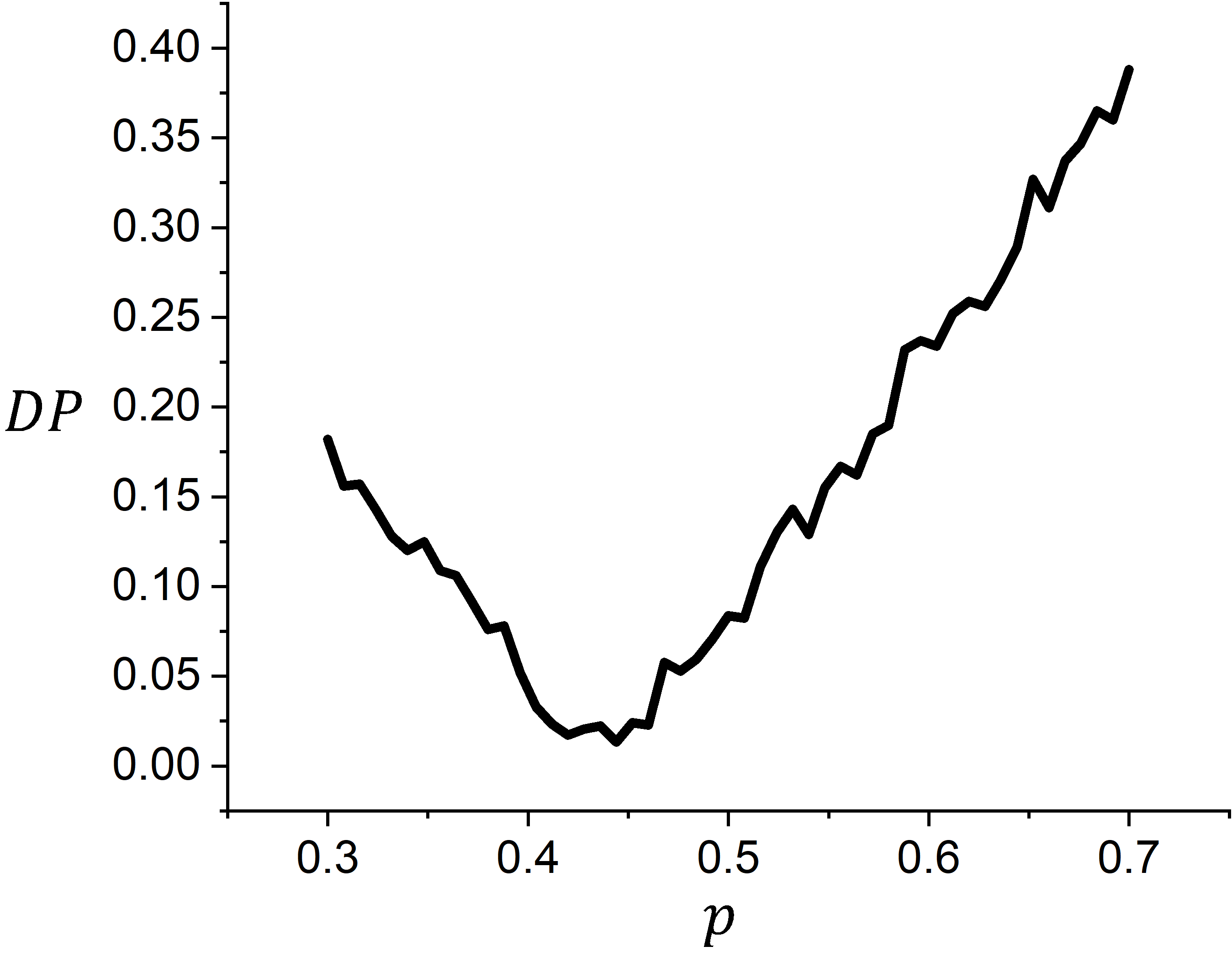}
    }
    
    \caption{The intercept planes in Figure \ref{fig:dif_bias_3D} at $p$ = 0.4 and 0.6, and $w_s$ = -1.0 and 1.0, respectively.}
    \label{fig:dif_bias_3D_intercept}
\end{figure}

Furthermore, unlike the real-world dataset, on the synthetic dataset, since we have ground truth, we can actually control which group the dataset is biased against as well as the degree of bias by adjusting the parameter $p$. Specifically, when the parameter $p$ is 0.5, the generated dataset is an unbiased dataset, while the parameter $p$ greater or less than 0.5 represents that it favors the advantaged group “1" or disadvantaged group “0", respectively. As such, we have the capability to construct datasets manifesting varying degrees of bias by manipulating the parameter $p$. On this basis, through the previously described method, controlling the parameter $w_s$ can construct ML models' decision processes with various biases. This enables further investigation of the impact on the ML model's distributive fairness when the dataset and the ML model's decision process have various degrees of same and different biases.  Specifically, we established a parameter range for $p$ within [0.3, 0.7], while maintaining $w_s$ within the range of [-5, 5]. From each of the above two ranges, we uniformly sampled 50 parameters, yielding a total of \num[group-separator={,}]{2500} combinations. We visualized the value of the \textit{DP} metric in each case, as shown in Figure \ref{fig:dif_bias_3D}. Further, to directly visualize the trend of the \textit{DP} metric with the $w_s$ and $p$ parameters, we intercepted the intercept planes for $p$ = 0.4 and 0.6, and $w_s$ = -1.0 and 1.0, respectively, as shown in Figure \ref{fig:dif_bias_3D_intercept}.

From Figure \ref{fig:dif_bias_3D_intercept}, we can see that consistent with previous findings, when the dataset and the ML model's decision process have the same bias, the biases are superimposed on each other. When the two exhibit opposite biases, the biases cancel out or even balance each other. Figure \ref{fig:dif_bias_3D_intercept_a} is used as an example to illustrate this in detail. Focusing on the intercepted plane with $p$ = 0.4, indicating dataset favoritism toward the disadvantaged group “0". When $w_s$ falls below 0.0, signifying that the ML model's decision process similarly favors the disadvantaged group “0", the \textit{DP} metric increases with the increasing absolute value of $w_s$. In contrast, when $w_s$ exceeds 0.0, implying the ML model's decision process preference for the advantaged group “1", the \textit{DP} metric initially decreases (indicating that the biases in the ML model's decision process and the bias inherent in the dataset gradually cancel out) before subsequently rising (signifying that the bias in the ML model's decision process exceeds the bias inherent in the dataset).

In addition, the results of this experiment further validate the conclusions in \textit{RQ1.1---RQ1.3}. When the parameter $p$ is 0.5 and the parameter $w_s$ is 0.0, the \textit{DP} metric value is very small as 0.013 (from Figure \ref{fig:dif_bias_3D}, not shown separately), which is consistent with the corresponding results in Table \ref{tab:fairData-fairModel}, confirming the conclusions in \textit{RQ1.1}. Whereas, as shown in Figs. \ref{fig:dif_bias_3D_intercept_a} and \ref{fig:dif_bias_3D_intercept_b}, the \textit{DP} metric value is very large when the parameter $w_s$ is 0.0, which implies that the inherent bias of the dataset leads to distributive unfairness of the ML model, corroborating the conclusions in \textit{RQ1.2}. Similarly, as shown in Figs. \ref{fig:dif_bias_3D_intercept_c} and \ref{fig:dif_bias_3D_intercept_d}, when the parameter $p$ is 0.5, the \textit{DP} metric value is also very large, which implies that the unfairness of the ML model's procedural unfairness affects its distributive fairness, corroborating the conclusions in \textit{RQ1.3}.

\begin{figure}[htpb]
    \centering
    \includegraphics[width=0.6\linewidth]{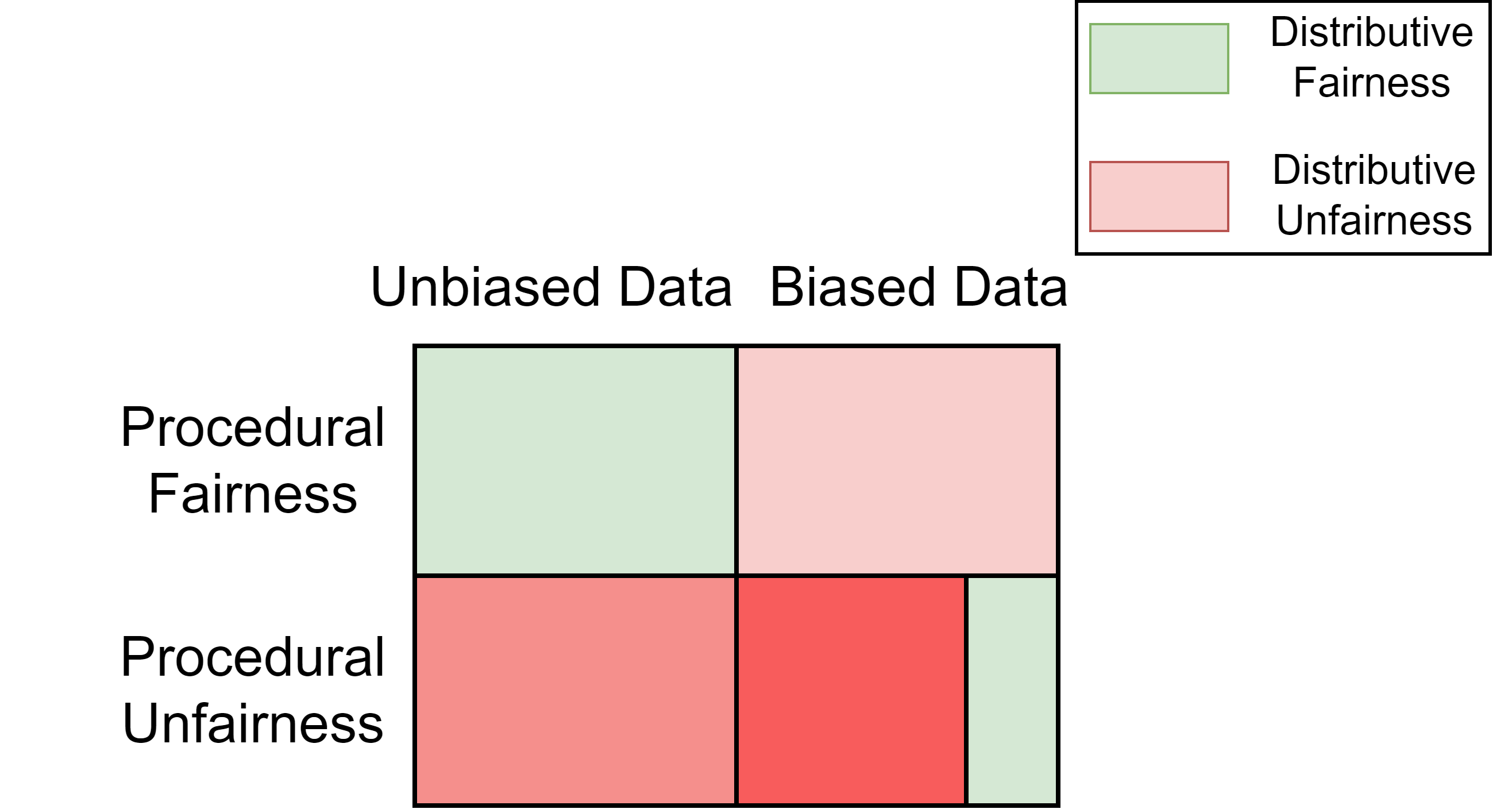}
    \caption{Impact of dataset bias and the ML model's procedural fairness on its distributive fairness. The depth of the red color represents the degree of distributive unfairness.}
    \label{fig:relationship}
\end{figure}

Overall, the impact of dataset bias and the ML model's procedural fairness on its distributive fairness is summarized in Figure \ref{fig:relationship}. When the dataset is unbiased and the ML model is procedurally-fair, it is also distributively-fair. Whereas either dataset bias or the ML model's procedural unfairness leads to the ML model's distributive unfairness, with the latter exerting a more pronounced influence. When both are biased, if both show the same bias, the bias will be superimposed to make the ML model's decision outcomes more unfair. On the contrary, the two will cancel each other out and even make the ML model exhibit distributive fairness. 

In summary, the above findings help us to clarify and even trace the source of the ML model's distributive unfairness. On one hand, the above experimental results illustrate that the ML model's distributive fairness stems precisely from the inherent bias of the dataset and the ML model's procedural unfairness. On the other hand, when we find that an ML model exhibits distributive unfairness, if it is procedurally-fair, we can know that the unfairness comes from the inherent bias in the dataset. Conversely, if we find that the dataset is unbiased, then the ML model's distributive unfairness stems from the ML model's decision-making process.

\subsection{RQ2 What Are the Differences Between Optimizing Procedural Fairness Metrics and Distributive Fairness Metrics in ML?}

After we have sorted out the impact of dataset bias and the ML model's procedural fairness on its distributive fairness, in this section we examine the difference between optimizing procedural and distributive fairness metrics. This is to gain a deeper understanding of the difference between the two dimensions in ML clearly and to provide guidance for people to choose the appropriate metrics and methods.

\textit{RQ2.1 How Does Optimization of Procedural Fairness Metrics Affect the Fairness of the ML Model?}

From the experimental results presented in Section \ref{sec:method}, we can conclude that optimizing the procedural fairness metric \textit{GPF\textsubscript{FAE}} yields a significant improvement in both procedural and distributive fairness within the ML model. This phenomenon can be explained by the conclusions drawn from our exploration of \textit{RQ1}. In \textit{RQ1}, we established that the ML model's distributive unfairness originates from the inherent bias within the dataset and the unfairness of the ML model's decision-making process. Consequently, optimizing the \textit{GPF\textsubscript{FAE}} metric and thus ensuring a fair decision-making process, means that some of the sources of distributive unfairness in the ML model are eliminated at the root, leading to a significant improvement in the distributive fairness of the ML model. Overall, optimizing the procedural fairness metrics can effectively mitigate and even eliminate the biases introduced or magnified by the ML model, which tends to be equally beneficial to the ML model's distributive fairness.

\textit{RQ2.2 How Does Optimization of Distributive Fairness Metrics Affect the Fairness of the ML Model?}

To address this sub-question, we compared the changes in procedural fairness and distributive fairness of ML models before and after considering the distributive fairness metric on \textbf{biased datasets}. The baseline is again the normally trained \textit{MLP\textsubscript{BCE}} model, i.e., the corresponding results of the \textit{MLP\textsubscript{BCE}} model in Table \ref{tab:method_compare}. We used three in-process methods for optimizing the distributive fairness metric \textit{DP}, i.e., regularization-based~\citep{kamishima2012fairness}, constraint-based~\citep{agarwal2018reductions}, and adversarial learning-based~\citep{zhang2018mitigating} methods, to examine the impact of optimizing the distributive fairness metric on the fairness of the ML model.

\begin{figure}[htpb]
    \centering
    \subfigure{
		\includegraphics[width=0.45\linewidth]{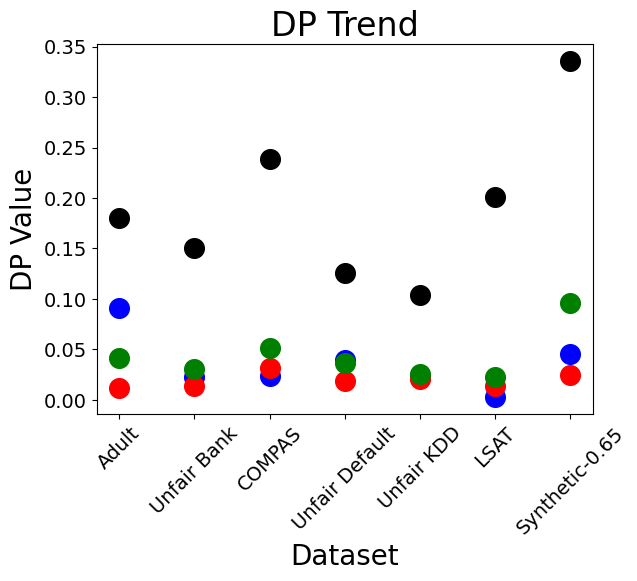}}
    \subfigure{
		\includegraphics[width=0.45\linewidth]{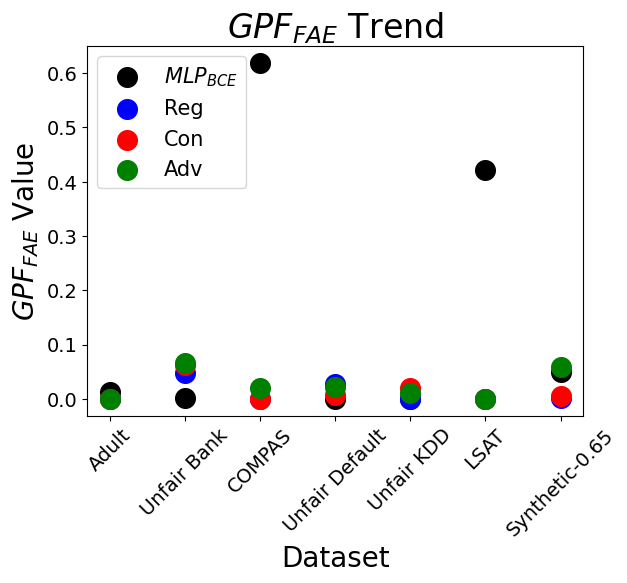}}
    \caption{Trends in the distributive fairness metric \textit{DP} and the procedural fairness metric \textit{GPF\textsubscript{FAE}} before and after considering the distributive fairness metric \textit{DP}.}
    \label{fig:compare_opt_DP}
\end{figure}

In the regularization-based method~\citep{kamishima2012fairness}, the hyper-parameter $\alpha$ for weighing BCE and \textit{DP} metrics was set to 0.2, 1.0, 0.2, 0.2, 0.2, 0.5, 0.2, and 0.4 on each dataset, respectively. On the other hand, for the constraint-based~\citep{agarwal2018reductions} and adversarial learning-based~\citep{zhang2018mitigating} methods, we used the implementations provided by the open-source Python algorithmic fairness toolkit, Fairlearn~\citep{bird2020fairlearn}, maintaining the default parameter configurations. It is worth noting that our focus is on exploring the impact of considering the \textit{DP} metric. Consequently, we have not deliberately fine-tuned the hyper-parameters associated with the above three methods on each dataset to pursue extreme performance, as typical hyper-parameter values suffice. The results of the comparison of the distributive fairness metric \textit{DP} and procedural fairness metric \textit{GPF\textsubscript{FAE}} are shown in Figure \ref{fig:compare_opt_DP}.

Analyzing Figure \ref{fig:compare_opt_DP}, it becomes evident that optimizing the \textit{DP} metric can significantly reduce the \textit{DP} metric value, which means that the distributive fairness is significantly improved. However, the \textit{GPF\textsubscript{FAE}} metric remains close to 0.0, which means that the ML model's decision-making process is still very unfair. Even on \textit{COMPAS} and \textit{LSAT}, two datasets where the decision-making process is originally relatively fair, the decision-making process becomes unfair instead after optimizing the \textit{DP} metric. This observation underscores a key implication: optimizing the distributive fairness metric contributes to fair decision outcomes, yet the decision-making process may either remain unfair or transition towards unfairness.

\begin{figure}[t]
    \centering
  \subfigure[Adult]{
		\includegraphics[width=0.45\linewidth]{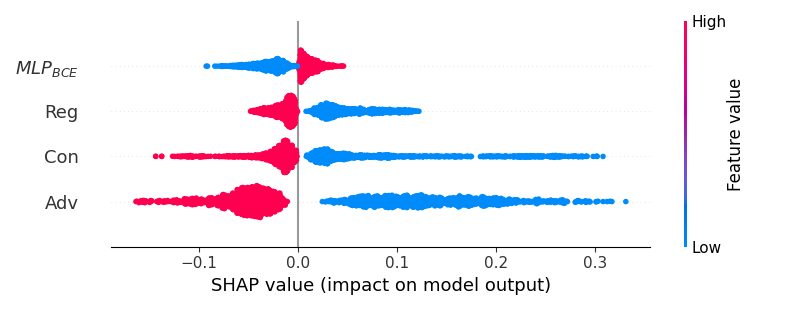}}
  \subfigure[Unfair Bank]{
		\includegraphics[width=0.45\linewidth]{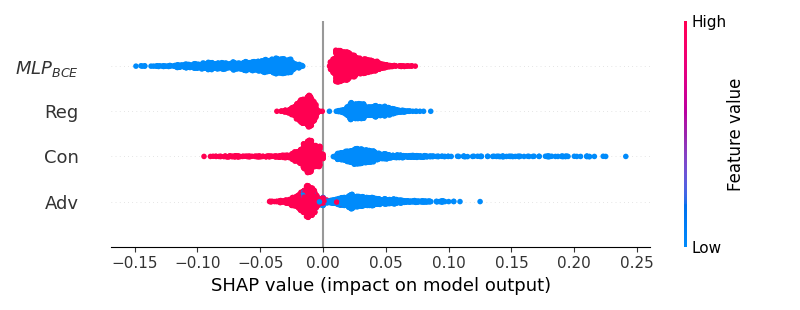}}
  \subfigure[COMPAS]{
		\includegraphics[width=0.45\linewidth]{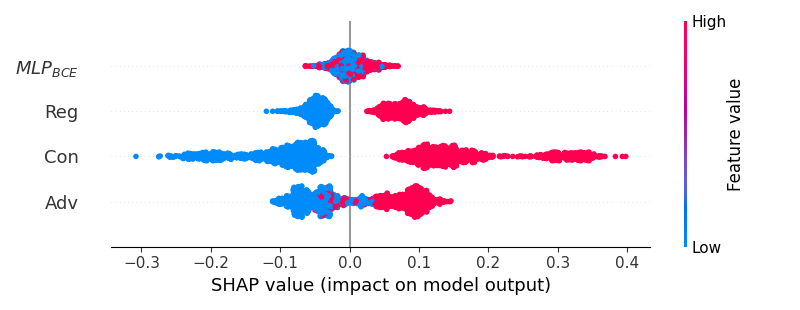}}
  \subfigure[Unfair Default]{
		\includegraphics[width=0.45\linewidth]{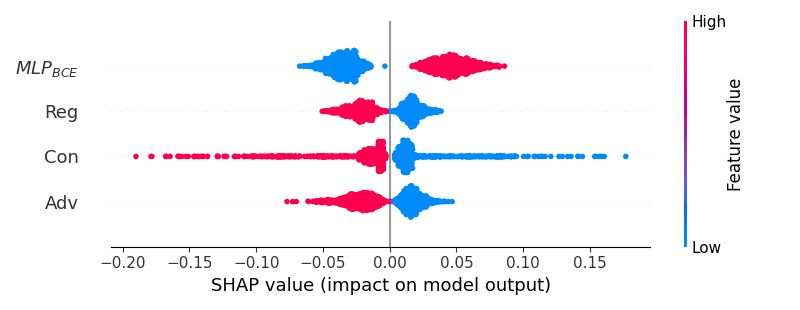}}
  \subfigure[Unfair KDD]{
		\includegraphics[width=0.45\linewidth]{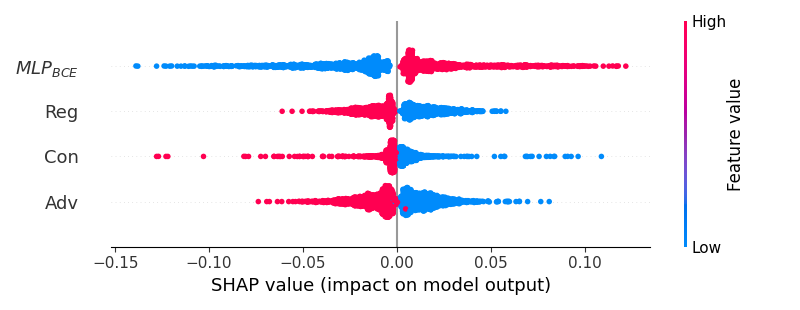}}
  \subfigure[LSAT]{
		\includegraphics[width=0.45\linewidth]{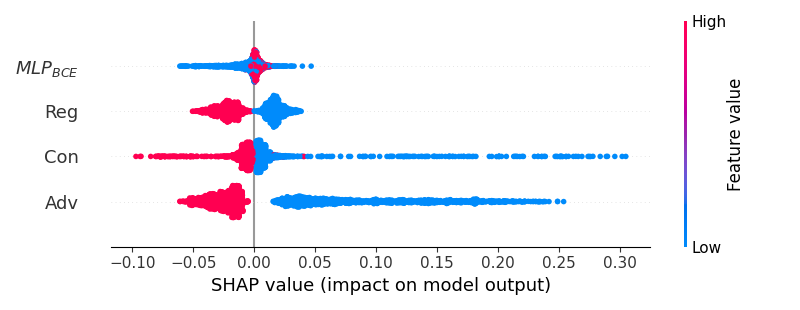}}
  \subfigure[Synthetic-0.65]{
		\includegraphics[width=0.45\linewidth]{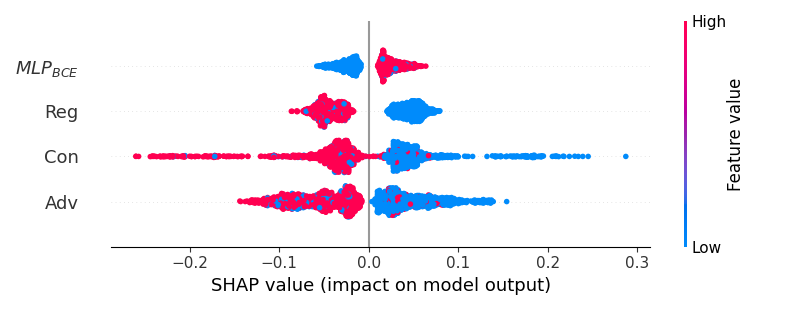}}
    \caption{Comparing the \textit{SHAP} value of the sensitive attribute in the ML model before and after optimizing the distributive fairness metric \textit{DP} on each dataset. Each red and blue dot represents a data point for an advantaged and disadvantaged group, respectively. The horizontal coordinate values are the results of the \textit{SHAP} method's explanation of the sensitive attribute of the corresponding data point. Larger absolute \textit{SHAP} values represent a greater impact on the decision, while a positive/negative sign indicates a positive/negative benefit to the decision.}
    \label{fig:opt_dp_explain}
\end{figure}

To further examine the impact of optimizing the \textit{DP} metric on the ML model's procedural fairness, we similarly visualized the distribution of the \textit{SHAP} value of the sensitive attribute in the set $\textit{\textbf{X}}^{'}$ before and after considering the \textit{DP} metric, as shown in Figure \ref{fig:opt_dp_explain}. Again, each red and blue dot represents a data point for the advantaged and disadvantaged groups, respectively.

From Figure \ref{fig:opt_dp_explain}, we can see that when training an \textit{MLP\textsubscript{BCE}} model normally, it exhibits either a lack of a clear preference for the sensitive attribute (as observed in the \textit{COMPAS} and \textit{LSAT} datasets) or a tendency to favor the advantaged group (in other datasets). However, after considering the \textit{DP} metric, all three methods show a preference for the disadvantaged group on the sensitive attribute. It is worth noting that the distribution of explanations for the sensitive attribute in the \textit{COMPAS} dataset in Figure \ref{fig:opt_dp_explain} is exactly the opposite of the other datasets. This is because the task of the \textit{COMPAS} dataset is to predict whether or not a person will reoffend, and predicting the positive category (i.e., that there will be recidivism) is instead unfavorable, and thus the distribution of its explanations is exactly the opposite, which precisely means that the ML model is unfavorable to the advantaged group.

This trend sharply contrasts with the phenomenon observed in Figure \ref{fig:mlp_vs_pf_explain}, where the sensitive attribute (almost) no longer has an impact on the decision after optimizing the procedural fairness loss $\mathcal{L}_{GPF}$. In contrast, after optimizing the \textit{DP} metric, the decision-making process of the ML model shifts from favoring the advantaged group or having no clear preference to favoring the disadvantaged group, when the distributive fairness metrics are significantly lower. This is consistent with the phenomenon that when the two show opposite biases in our \textit{RQ1.4}, the biases cancel each other out or even appear in a balanced state making the ML model achieve distributive fairness. Overall, optimizing the distributive fairness metric leads to their use of the bias of the ML model's decision-making process toward the disadvantaged group to offset the data's inherent bias toward the advantaged group, thus achieving distributive fairness of the ML model.

\begin{figure}[htpb]
    \centering
    \includegraphics[width=0.6\linewidth]{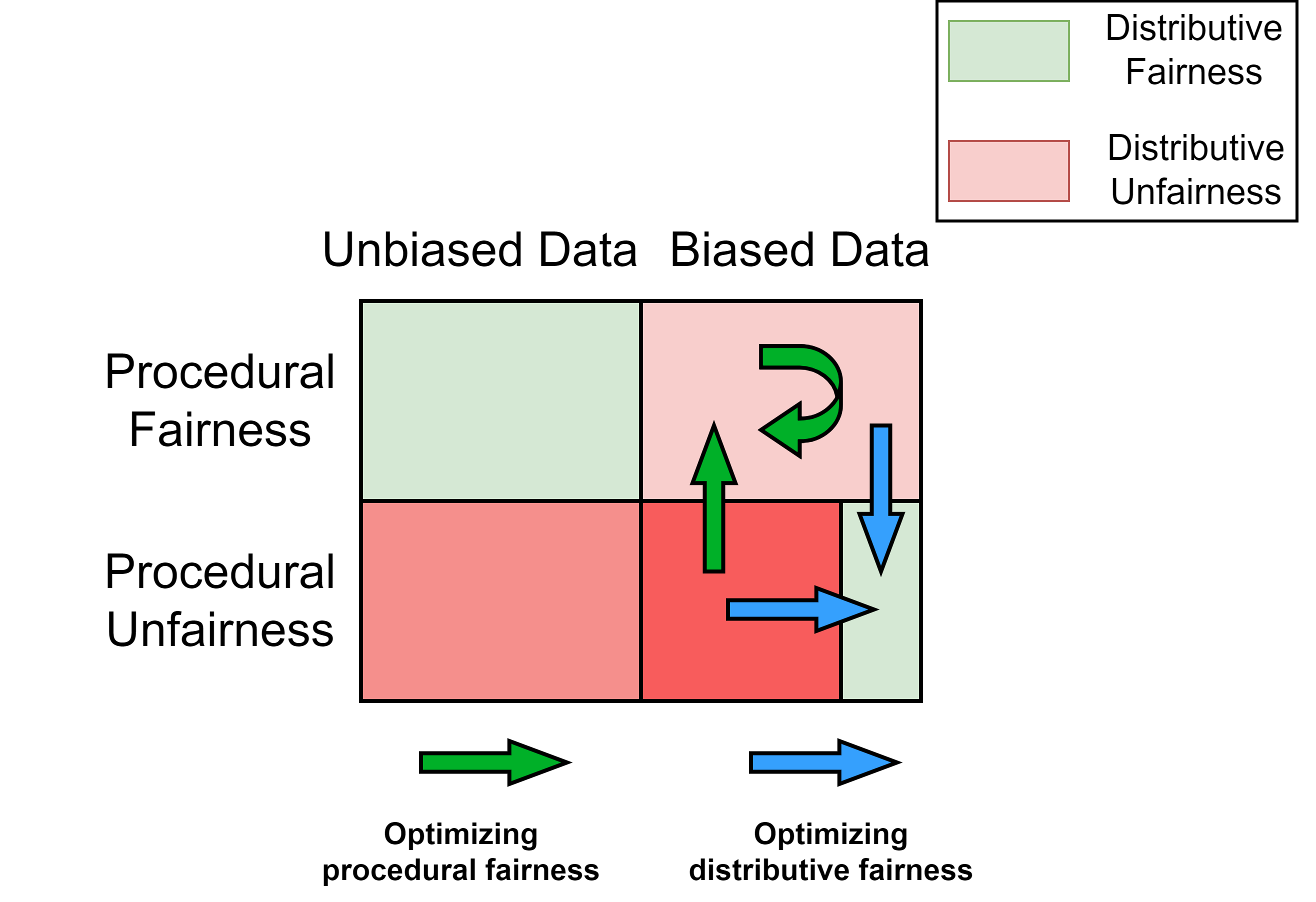}
    \caption{Comparison of the difference between optimizing procedural fairness and distributive fairness in ML.}
    \label{fig:transformation_relationship}
\end{figure}

In summary, the optimization of the procedural fairness metric and distributive fairness metric entails distinctly divergent outcomes. Optimizing the procedural fairness metric serves to rectify biases introduced or magnified by the ML model's decision process, and improves the ML model's distributive fairness alongside procedural fairness. Nonetheless, at this juncture, the ML model's decision outcomes may still exhibit unfairness, stemming from the inherent biases embedded within the dataset. Conversely, optimizing the distributive fairness metric will obtain an ML model in which the decision-making process favors the disadvantaged group to counteract the bias inherent in the dataset. This approach yields an ML model with fair decision outcomes, yet it remains inherently unfair in terms of both the underlying data and the ML model's decision-making process. The distinction between the two optimization strategies is visually encapsulated in Figure \ref{fig:transformation_relationship}.

\section{Discussions}
\label{sec:discussion}

After we have elucidated the relationship between the ML model's procedural fairness and distributive fairness, we engage in further discussion in this section. It includes considerations of trade-offs between the two dimensions, suggests a new path of achieving the ML model's distributive fairness as well as discusses the impact of different distributive fairness metrics on our conclusions.

\subsection{Trade-off between the ML Model's Procedural Fairness and Distributive Fairness}

Previous literature has highlighted the existence of trade-offs between model accuracy with distributive fairness~\citep{caton2020fairness,friedler2019comparative,speicher2018unified} and procedural fairness~\citep{wang2024procedural}, respectively. Additionally,~\cite{wang2024procedural} have also observed the existence of a complex relationship between the ML model's procedural fairness and distributive fairness, which are sometimes consistent and sometimes have trade-offs and conflicts.

Our study provides insights that offer an explanation for this phenomenon. The trade-off between the two dimensions stems precisely from the bias inherent in the dataset. The ML model's procedural fairness primarily aims to ensure that the ML model's internal decision-making process is fair. In contrast, the ML model's distributive fairness is needed to counteract the dataset's inherent preference for advantaged groups by using the decision-making process's preference for the disadvantaged group. As a result, when the ML model exhibits pronounced unfairness, irrespective of whether procedural fairness or distributive fairness metrics are prioritized, both dimensions of fairness tend to improve, at which point the two perform consistently.

However, in scenarios where the decision-making process of the ML model is already fair, continued optimization of distributive fairness metrics necessitates a trade-off, often at the expense of procedural fairness. This trade-off is driven by the imperative to offset the dataset's inherent bias, ultimately achieving distributive fairness. Similarly, continuing to optimize procedural fairness when the ML model outcomes are fair leads to distributive unfairness, which is caused by the inherent bias of the dataset.

\subsection{A New Path for Achieving Distributive Fairness in ML}

In the ML models, optimizing distributive fairness metrics and thus using the ML model's decision-making process to favor disadvantaged groups to offset the inherent bias of the dataset may raise valid concerns. This pursuit of distributive fairness, achieved at the potential cost of sacrificing procedural fairness, invites scrutiny.

Contrastingly, our experimental findings highlight an alternative path to achieving distributive fairness. From \textit{RQ1}, we know that when the dataset remains unbiased, and the ML model's decision-making process is fair, the resulting ML model yields fair decision outcomes. Therefore, we can obtain a distributively-fair ML model by ensuring that the data are unbiased and the ML model is procedurally-fair, thus curbing unfairness at the source. The potential road is to ensure that the dataset is unbiased through some pre-process technique~\citep{zemel2013learning,kamiran2012data}, while achieving procedural fairness through, for example, the approach we propose in Section \ref{sec:method}. We believe this is a more desirable way to achieve distributive fairness, unlike pursuing distributive fairness itself, but rather focusing on mitigating unfairness at its roots. Crucially, both the data and the ML model's decision-making process are fair at this point.

\subsection{Impact of Different Distributive Fairness Metrics on Conclusions}

Nowadays, numerous distributive fairness metrics have been proposed, and they focus on different aspects of the ML model's distributive fairness~\citep{9902997}. In this paper, we have chosen to center our investigation on the \textit{DP} metric, primarily due to its widespread adoption and recognition as a distributive fairness metric~\citep{jiang2021generalized}. Given the multitude of distributive fairness metrics available, individually examining their relationships with the ML model's procedural fairness would be impractical. Nevertheless, to ensure both the accuracy and applicability of our findings, we also made related attempts on two other widely recognized fairness metrics, equal opportunity (EOP)~\citep{hardt2016equality} and equalized odds (EOD)~\citep{hardt2016equality}, and obtained similar conclusions. Consequently, we assert that the variance in distributive fairness metrics does not significantly impact the conclusions presented in this paper.

\section{Conclusion}
\label{sec:conclusion}

In this paper, we propose a method for achieving procedural fairness during training ML models. Its effectiveness is verified on a synthetic and six real-world datasets. Furthermore, we explore and examine the relationship between the ML model's procedural fairness and distributive fairness with the help of our proposed method. On one hand, we investigate the impact of the inherent bias of the dataset and the ML model's procedural fairness on its distributive fairness. The results show that (1) when the dataset is unbiased and the ML model is procedurally-fair, it is also distributively-fair; (2) either the inherent bias of the dataset or the unfairness of the ML model's decision-making process affects the fairness of the ML model's decision outcomes, with the latter exerting a more pronounced impact; and (3) when the dataset and the ML model's decision-making process are unfair, the effects of the two on distributive fairness mutually superimposed or canceled out, depending on whether the biases of the two are consistent or opposite. On the other hand, we investigate the difference between optimizing procedural and distributive fairness metrics in ML. Optimizing procedural fairness metrics helps eliminate biases introduced or amplified by the ML model, ensuring a fair decision-making process. In contrast, optimizing distributive fairness metrics results in an ML model with a decision-making process that favors the disadvantaged group, leveraging this preference to counterbalance the inherent dataset bias in favor of the advantaged group, thus achieving distributive fairness.

In the future, we plan to use multi-objective optimization algorithms~\citep{chandra2006ensemble} to obtain models with different trade-offs in model accuracy, procedural fairness, and distributive fairness.

\acks{This work was supported by the National Natural Science Foundation of China (Grant No. 62250710682), the Guangdong Provincial Key Laboratory (Grant No. 2020B121201001), and the Program for Guangdong Introducing Innovative and Entrepreneurial Teams (Grant No.2017ZT07X386).}

\end{document}